\documentclass[runningheads]{llncs}
\usepackage{graphicx}

\usepackage{multirow}

\usepackage{url}
\usepackage[colorlinks=true,linkcolor=black,urlcolor=blue,citecolor=black]{hyperref}

\usepackage{amsmath}
\usepackage{amssymb}
\usepackage{amsfonts}

\usepackage{enumitem}

\usepackage{xcolor}

\usepackage[ruled,vlined]{algorithm2e}
\begin{document}
\title{Template-Based Graph Clustering\thanks{Supported by Agence Nationale de la Recherche (ANR), project ANR-17-CE23-0021, and São Paulo Research Foundation (FAPESP), project 2017/50236-1. F. Yger acknowledges the support of the ANR as part of the ``Investissements d'avenir" program, reference ANR-19-P3IA-0001 (PRAIRIE 3IA Institute).}}
%
%
\author{Mateus Riva\inst{1,2} \and
Florian Yger\inst{2}\and
Pietro Gori\inst{1} \and
Roberto M. Cesar Jr.\inst{3} \and
Isabelle Bloch\inst{1}}
\authorrunning{M. Riva et al.}
%
\institute{LTCI, Télécom Paris, Institut Polytechnique de Paris, France\\
\email{\{mateus.riva, pietro.gori, isabelle.bloch\}@telecom-paris.fr} \and
LAMSADE, Université Paris-Dauphine, PSL Research University,  France\\
\email{florian.yger@dauphine.fr} \and
IME, University of São Paulo, Brazil\\
\email{rmcesar@usp.br}}
\maketitle              
\begin{abstract}
We propose a novel graph clustering method guided by additional information on the underlying structure of the clusters (or communities). The problem is formulated as the matching of a graph to a template with smaller dimension, hence matching $n$ vertices of the observed graph (to be clustered) to the $k$ vertices of a template graph, using its edges as support information, and relaxed on the set of orthonormal matrices in order to find a $k$ dimensional embedding. 
With relevant priors that encode the density of the clusters and their relationships, our method outperforms classical methods, especially for challenging cases.

\end{abstract}
%
%
%

\section{Introduction}
\label{sec:introduction}

\begin{figure}[t]
    \centering
    \includegraphics[width=0.9\textwidth]{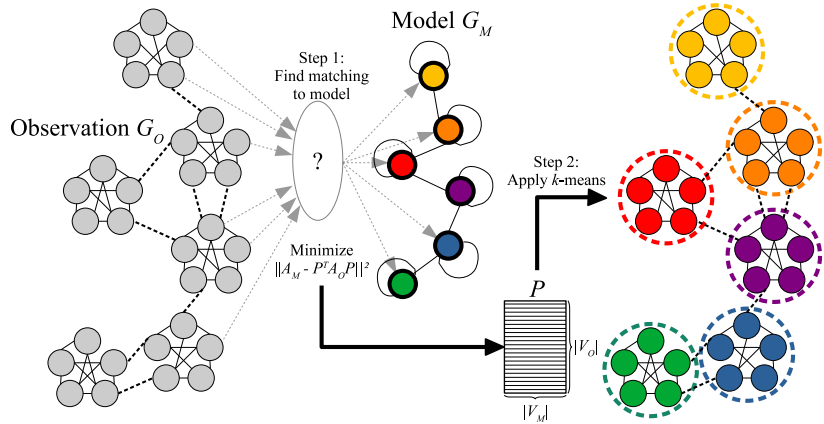}
    \caption{Visual summary of the proposed template-based graph clustering approach. Given an observation graph $G_O$ (on the left) and a model graph $G_M$ representing the underlying communities and their relationships, we match every vertex in $G_O$ to a single vertex in $G_M$. This assignment is represented as a transformation matrix $P$, which acts as an embedding of the vertices in $G_O$, and optimized so as to minimize an objective function based of the adjacency 
    matrices of the graphs. We apply $k$-means clustering to $P$ to find the final clusters.}
    \label{fig:graphical_abstract}
\end{figure}

Graph clustering consists in labeling vertices of a graph (or network) in a way that vertices with the same label belong to a set (alternatively, cluster or community) with some sort of coherent relation. Typically, graph clustering is unsupervised, that is, there is no prior information on the shape and nature of the sets to which vertices should be clustered into. However, in some applications,  prior information on the underlying structure of the clusters of the data may exist. This can be modeled as a ``template'' (or model) graph where each vertex corresponds to a different cluster. When this information is available, we would like to make use of it to improve the quality of a clustering.

A common example of this scenario are graphs extracted from segmentations of structured scenes. 
In a structured scene, the objects are spatially arranged according to a known structure. For example in medical images, organs in the body are organized in a fixed anatomical structure.
This prior encodes a significant amount of information that can help in segmenting the objects in the image; by clustering a graph 
built from image information with a template of the underlying structure, better results could be achieved.

To this end, we propose a matching-based graph clustering technique. Each vertex of an ``observed'' graph (the one to be clustered) is matched to a single vertex of the ``model'' graph. Based on this mapping, the label of the clusters are then transferred from the ``model'' to the  ``observed'' graph. This approach allows us to leverage prior information on the underlying structure of the communities by encoding this information in the model graph.


The main contribution of this paper is a novel method for the clustering of graphs into communities that obey a known prior underlying structure. 
The proposed approach is summarized in Figure~\ref{fig:graphical_abstract}.

\subsection*{Definitions and notations}
\label{ssec:introduction_definitions}

Let $G = (V,E)$ be a graph, where $V$ is the set of vertices and $E$ the set of edges. Vertices are denoted by $v_i$ or simply $i$, $i=1...|V|$, and edges are denoted by $e_{ij}$ where $i$ and $j$ denote the indices of the vertices of the edge. If the edges are weighted, then the weight of $e_{ij}$ is denoted by $w_{ij}$.
Let $A\in \mathbb{R}^{|V|\times |V|}$ be the graph adjacency matrix, where an element $a_{ij}$ of $A$ is equal to $w_{ij}$ if there is an edge of weight $w_{ij}$ in $E$ between vertices $v_i$ and $v_j$, and 0 otherwise. Let $D \in \mathbb{R}^{|V|\times|V|}$ be the graph degree matrix, a diagonal matrix where $d_{ii}= deg(v_i)$, where $deg(v_i)$ denotes the degree of vertex $v_i$.
We define the Laplacian $L$ of a graph, $L \in \mathbb{R}^{|V|\times|V|}$, as $L=D-A$.\footnote{Note that other definitions of the Laplacian exist~\cite{Luxburg2007-SC}.}

In the proposed approach, graph clustering is driven by a model, and the problem is then formulated as follows. We define $G_M$ as a model, or template, graph containing $|V_M|=k$ vertices, where $A_M$ is its adjacency matrix and $L_M$ its Laplacian. The model graph represents the expected basic structure of the data, where each vertex represents a single cluster, and thus $k$ should be equal to the number of clusters. We then define $G_O$ as an observation graph containing $|V_O|=n$ vertices, where $n > k$, $A_O$ is its adjacency matrix and $L_O$ its Laplacian. The observation graph represents the data to be clustered. The clustering is expressed as a matching problem, where each vertex in $G_O$ is matched to a single vertex in $G_M$, hence a cluster (or community) is a set of vertices of $G_O$ matched to the same vertex of $G_M$.

\section{Related Work}
\label{sec:related_work}

\paragraph{Spectral Graph Clustering.}
\label{ssec:spectral_graph_clustering}
Spectral Graph Clustering~\cite{Luxburg2007-SC} is a popular technique for clustering data organized as graphs. Data can be either originally represented as a graph, or similarity graphs can be built from the data. 
From the Laplacian $L$ of a graph, its first $k$ eigenvectors are computed (i.e. the ones associated with the smallest eigenvalues). The eigenvectors are then stacked as columns of a matrix where each row $i$ would encode a $k$-dimensional embedding of the node $i$ from the original graph. Then, $k$-means clustering is applied on this embedding.

An application of Spectral Clustering~\cite{Shi2000-PAMI} was proposed for solving perceptual grouping in images as a normalized version of Ratio-Cut~\cite{Hagen1992-CADICS}, thus introducing spectral clustering to the machine learning community. Apart from the normalized-cut approach proposed by Shi and Malik, Ng et al. \cite{Ng2002-NIPS} proposed to normalize the rows of the embedding to unit length before applying $k$-means. Further variations and extensions of the method can be found in \cite{Nascimento2011-EJOR,Tolic2018-PR}.

Typical definitions of spectral clustering are unable to exploit any form of prior information on the underlying structure of the communities beyond the total number $k$ of communities. Additionally, due to its spectral nature, the technique has a specific 
definition of ``community'' as being a set of nodes with high internal (intra-cluster) connectivity (i.e. between nodes of the same set) and low external (inter-cluster) connectivity (i.e. between nodes of different sets) -- a definition that fails to describe some scenarios, e.g. bipartite graphs.

\paragraph{Modularity Graph Clustering.}
Modularity $Q$~\cite{newman_modularity_2006,Newman2004-PRE} is a quality function to measure the community structure of a graph (i.e. clustering).
It is defined as:
\begin{equation}\label{eq:modularity}
Q = \frac{1}{2|E|}\sum_{i=1}^{|V|} \sum_{j=1}^{|V|}\left[a_{ij} - \frac{d_{ii}d_{jj}}{2|E|}\right]\delta(c_i, c_j)
\end{equation}
where $c_i$ is the cluster to which vertex $i$ belongs, and $\delta(c_i, c_j)$ is equal to $1$ if $c_i = c_j$ and $0$ otherwise.
Maximizing $Q$ amounts to search for $k'$ clusters with a high number of intra-cluster connections 
and a small number of inter-cluster connections.
It has been demonstrated that exact modularity optimization is strongly NP-complete~\cite{Blondel2008-SMTE,Clauset2004-PRW}.
The Clauset-Newman-Moore (CNM) algorithm~\cite{Clauset2004-PRW} for clustering a graph performs greedy modularity maximization. Each node is initially set as its own community; at each iteration, the pair of communities that most increases modularity is joined until no potential pair increases modularity. Other greedy modularity-based approaches have been proposed, such as the Louvain algorithm~\cite{Blondel2008-SMTE}, similar to CNM, but based on greedily changing a node label to that of its neighbor, and others~\cite{Brandes2007-KDE}.

It has been shown that modularity maximization algorithms have difficulty finding small communities in large networks and the resulting clusters tend to have similar sizes, as the expected number of inter-edges between communities gets smaller than one (the so-called resolution limit)~\cite{fortunato_resolution_2007}. Additionally, modularity maximization techniques automatically find a number of clusters $k'$, which can be seen as an advantage in some applications. However, in scenarios where we have prior information on the underlying structure of the clusters and access to the real number of clusters $k$, these techniques may not perform as well as those that incorporate this information. Finally, the definition of $Q$ implies the same specific definition of ``community'' as that of spectral clustering: a set of nodes with high internal connectivity and low external connectivity.






\section{Template-Based Graph Clustering}
\label{sec:tbgc}
In this section, we detail the proposed Template-Based (TB) graph clustering method.

\subsection{Basic Formulation of the Inexact Matching Problem}
\label{ssec:tbgc_matching}
We want to find a mapping $P$ which best maps vertices of $G_O$ to vertices of $G_M$. Many vertices of $G_O$ may be mapped to the same vertex in $G_M$, but a vertex of $G_O$ can only be mapped to a single vertex of $G_M$ (hence, \textit{many-to-one} matching). Similarly to 
\cite{Koutra2013-ICDM}, we will generally state the problem as solving:
\begin{equation}
\label{eq:matching_problem_adjacency}
  arg\,min_{P} F(P), \;  F(P) = \| A_M -  P^T A_O P \|^2_\mathcal{F}
\end{equation}
where $\| . \|_\mathcal{F}$ denotes the Frobenius norm, $P$ is a \textit{binary} transformation matrix such that $P \in \{0,1\}^{n \times k}$, and each line in $P$ (corresponding to one vertex of $G_O$) has exactly one non-zero value (corresponding to the associated vertex in $G_M$) such that $\sum_{j=1}^{k}p_{ij}=1, \, \forall i=1...n$, where each element $p_{ij}$ of the matrix $P$ indicates whether the observation vertex $i$ is part of the cluster $j$.


In this formulation, the $j$-th column of the matrix $P$ indicates vertices in the $j$-th cluster. The term $P^T A_O P$ contracts all the edges of the observation, representing the connections of each observed cluster as the edges of a single vertex in the model. It can be thought of as a compression of the $n \times n$ matrix $A_O$ into a $k \times k$ matrix that is similar to $A_M$, as can be seen in Figure~\ref{fig:graphical_abstract}.


The choice of the adjacency matrix instead of the Laplacian was due to the fact that the Laplacian of the model is unable to capture any information related to the size of each community. The adjacency matrix formulation, by contrast, encodes the quantity of edges inside a community as self-loops in the model.


\subsection{Approximating the Matching Problem}
\label{ssec:tbgc_approximation}
In the basic formulation of the solution, finding the optimal transformation matrix $P^*$ is a NP-hard problem. To find an approximation, we can relax the constraints on $P$ to $P \in \mathbb{R}^{n \times k}$ and $P^T P = \mathbb{I}_k$, where $\mathbb{I}_k$ denotes the identity matrix of size $k$, such that $P$ is now an orthonormal $k$-frame. Then instead of directly searching a matching, we seek an embedding for the vertices in a $k$-dimensional space, in the spirit of the embedding learned by Spectral Clustering. From this point forward, we will consider Equation~\ref{eq:matching_problem_adjacency} using this relaxed definition of $P$.

This constraint is smooth and defines a Riemannian manifold called the Stiefel manifold. Several algorithms exist for optimizing over this space~\cite{absil2009optimization,edelman1998geometry}. This allows us to perform a local search for an approximation of $P^*$  for function $F(P)$, which we call $P_{opt}$, using gradient descent on the Stiefel manifold.

On Riemannian manifolds, at every point a tangent plan (i.e. a linearization of the manifold equipped with a scalar product) is defined. The Euclidean gradient of a cost function at a point $X$ may not belong to the tangent space at $X$, and in such a case, the projection of the gradient on the tangent space makes it a Riemannian gradient. This Riemannian gradient parametrizes a curve on the manifold -- a geodesic -- that locally diminishes the cost function. For the sake of efficiency, an approximation -- called a retraction -- has been developed for the operation, transforming a displacement in the tangent space into a displacement in the manifold. This operation is denoted by $R_X(u)$ and is the transformation of a vector $u$ in the tangent space at the point $X$ into another point on the manifold. As for every gradient based algorithm, the choice of a step size is critical and line-search methods have been extended to this setup. For more details about optimization on matrix manifolds, the reader may want to refer to~\cite{absil2009optimization}.
We summarize the process of this search in Algorithm~\ref{alg:optimization}.

\begin{algorithm}[H]
\label{alg:optimization}
    initialize $P_0$ randomly\;
    \While{convergence is not satisfied}{
    $P_{t+1} \gets R_{P_t} \left( \eta_t \nabla_t (F)\right)$\;
    $t \gets t+1$
    }
    \KwResult{$P_t$}
    \caption{Search for optimized transformation $P_{opt}$ using Steepest Gradient Descent}
    
\end{algorithm}

Note that the Steepest Gradient Descent used in Algorithm~\ref{alg:optimization} is an optimization method that is vulnerable to local minima. Therefore, different initializations of $P_0$ may lead to different approximations of $P^*$. However, as the results in Section~\ref{sssec:synth_results} show, the approximations found from several distinct initializations produce stable results in practice.

The Euclidean gradient of $F$ defined in Equation \ref{eq:matching_problem_adjacency} is given by 
\begin{equation}
    \label{eq:gradient_adjacency}
    \frac{\partial F(P)}{\partial P} = 4(A_OPP^TA_OP-A_OPA_M)
\end{equation}
The equation for its projection on the tangent space (thus creating the Riemannian gradient $\nabla(F)$ in Algorithm~\ref{alg:optimization}) and for the retraction can be respectively found in~\cite[Ex~3.6.2~p.48]{absil2009optimization} and~\cite[Ex~4.1.3~p.59]{absil2009optimization}. The step size $\eta_t$ is found using line search.

\paragraph{$k$-Means Clustering.}




The optimized $P_{opt}$ can be seen as an analog to the eigenvector matrix in the context of Spectral Clustering: each row $i \in [1,|V_O|]$ in $P_{opt}$ corresponds to the $k$-dimensional embedding of the observation vertex $v_{O_i}$. If this embedding is of high quality, the distance, in this space, between points matched to the same model vertex should be small; we can then apply a clustering algorithm (such as $k$-means, which we use in this implementation) on $P_{opt}$. In this case, we would handle $P_{opt}$ as the input data matrix $X$, with each $x_i$ corresponding to a line of $P_{opt}$ and, consequently, an observation vertex.

Note that our current implementation utilizes a QR decomposition of the $P$ matrix for computing the retraction step, which has an approximate complexity of $\mathcal{O}(nk^2)$. Thus, scalability to larger datasets with a higher number of classes is beyond the scope of this current work, and limits the maximum size of experiments. However, there is space for further refinement of the optimization process, improving TB clustering scalability.

\section{Experiments}
\label{sec:experiments}

We conducted a series of experiments to demonstrate the effectiveness of the template-based graph clustering technique, and compare it with classic baselines such as Spectral Clustering~\cite{Luxburg2007-SC} and both Clauset-Newman-Moore (CNM)~\cite{Clauset2004-PRW} and Louvain~\cite{Blondel2008-SMTE} Modularity clustering (see Section~\ref{sec:related_work}). We performed proof-of-concept experiments on synthetic datasets (Section~\ref{ssec:synth_experiments}) and on real datasets (Section~\ref{ssec:real_experiments}).

The evaluation is performed according to the following measures:
\begin{itemize}[topsep=0pt]
    \item \textbf{Adjusted Rand Index}~\cite{Hubert1985-JC} (ARI) of the clustering, used as a measure of clustering accuracy and quality. The Rand Index is a measure of agreement between two partitions of a set, and can be computed as $RI = \frac{TP + TN}{TP+TN+FP+FN}$ (where $TP,TN,FP,FN$ are the numbers of true positives, true negatives, false positives and false negatives, respectively). The ARI is corrected-for-chance by using the expected similarity of comparisons as a baseline.
    This index takes values in $[-1,1]$, with 0 meaning a random clustering and 1 meaning a perfect clustering (up to a permutation);
    \item \textbf{Projector Distance}: as the matching of points to clusters can be thought as an embedding $P$, and as we are searching for an embedding that approaches the ``perfect'' embedding $P^*$ represented by the ground truth\footnote{In practice, $P^*$ is not always an orthogonal matrix and to be able to use this distance, we compute its closest orthogonal matrix.} (i.e. reference classification of the data), we compute the Frobenius norm of the difference between the projector of these embeddings: $PD = \|PP^T - {P^*}{P^*}^T\|^2_\mathcal{F}$. 
\end{itemize}

The Python code used to run all experiments along with the results generated are available online\footnote{\url{https://github.com/MarEe0/TBGC}}. For solving the manifold optimization involved in Equation \ref{eq:matching_problem_adjacency}, the toolbox Pymanopt~\cite{pymanopt} was used. For oeprations based on graphs, the NetworkX~\cite{Hagberg2008-networkx} and Scikit-Network\footnote{\url{https://scikit-network.readthedocs.io/}} libraries were used.

\subsection{Experiments on Synthetic Datasets}
\label{ssec:synth_experiments}
In order to provide a controlled environment to study in depth the performance of the TB technique, experiments were performed on different synthetic datasets. These datasets allowed us to easily control characteristics such as size of clusters and probabilities of inter- and intra-connection of communities. All synthetic datasets used are undirected and unweighted (all edges have weight $w=1$).

For each experiment, the following steps were taken:
\begin{enumerate}[topsep=0pt]
    \item Generate a random example of the specified toy dataset;
    \item Execute template-based, pure spectral, CNM modularity and Louvain modularity clustering methods;
    \item Compute the two evaluation measures (adjusted rand index and projector distance) from the output of the clustering;
    \item Perform 100 repetitions with different random initializations, compute average and standard deviation of the evaluation measures.
\end{enumerate}

\subsubsection{Description of Synthetic Datasets}

\paragraph{The $G_3$ graph:} a graph containing three communities connected in a line, i.e. cluster 1 is connected with cluster 2, which is connected to cluster 3. Probabilities of inter-connection are $0.1$ between communities 1 and 2 and between 2 and 3; probabilities of intra-connection are the complement of the inter-connection probabilities: $0.8$ for the central community and $0.9$ for the outer communities.

\paragraph{The $G_6$ graph:} a graph containing six communities with no model isomorphisms. Probabilities of inter-connection are $0.1$ between all communities; probabilities of intra-connection are the complement of the inter-connection probabilities: $1 - 0.1deg(community_i)$ for each community $i$, where $deg(community_i)$ is the number of other communities that have vertices directly connected to vertices belonging to community $i$.

\paragraph{The $C2$ graph:} a graph containing four communities in a line; in our experiments, we varied the inter-connection probability of the two central communities in order to test a progressively harder clustering scenario. Probabilities of inter-connection are $0.1$ between the outer and the central communities, and variable between the central communities. Intra-connection probability is the complement of the inter-connection probabilities.

\paragraph{The $BP$ graph:} a graph containing only two communities; used for experimenting on bipartite graphs (intra-connection probability equal to $0$ for all communities) and hub (or star)-style graphs (intra-connection probability equal to $0$ for one community). We varied the inter-connection probabilities.

Figure~\ref{fig:synth_examples} shows illustrative examples of all synthetic graphs used. We display the observation graphs, and the corresponding model graph is represented by colors (one color for each cluster or each vertex of the model graph).

{\setlength{\tabcolsep}{0.09in}
\begin{figure}[htbp]
    \centering
    \begin{tabular}{cccc}
        \multirow{3}{*}[0.8in]{\includegraphics[width=0.18\textwidth]{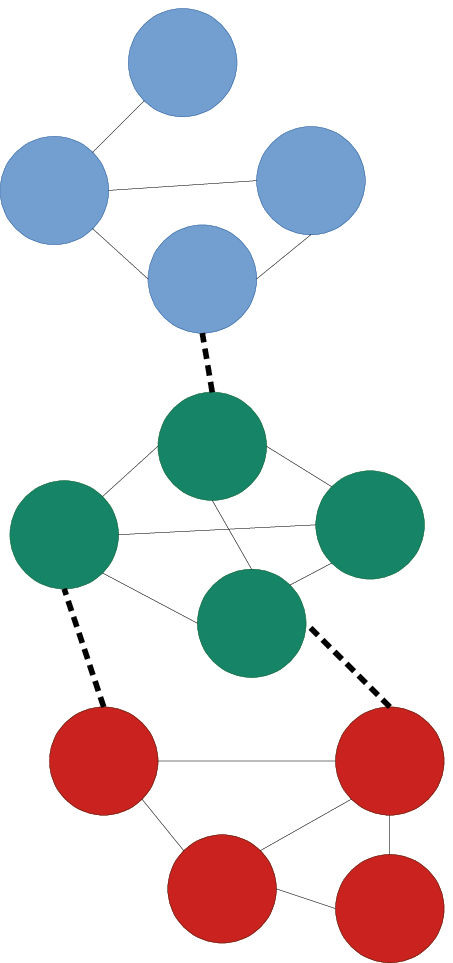}} & 
        \multirow{3}{*}[0.8in]{\includegraphics[width=0.2\textwidth]{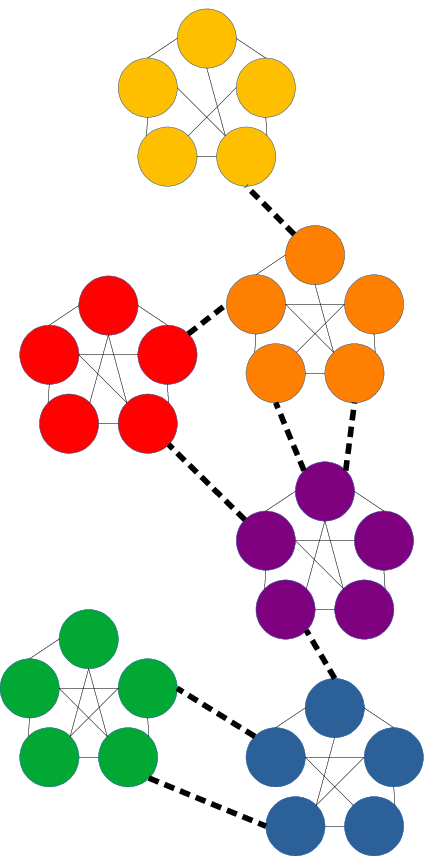}} & 
        \multirow{3}{*}[0.8in]{\includegraphics[width=0.145\textwidth]{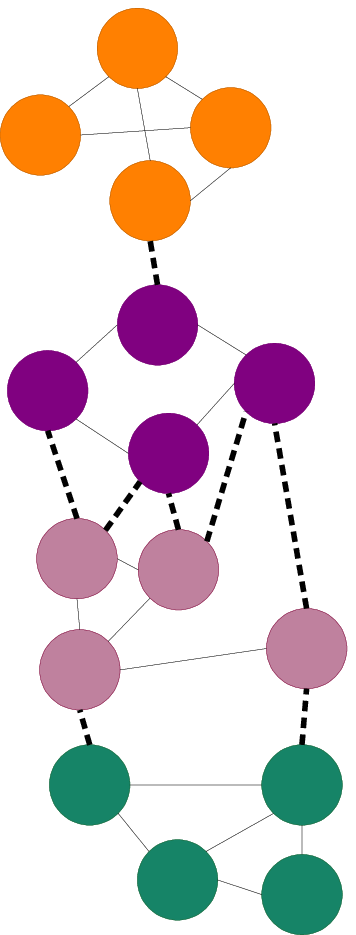}} & 
        \includegraphics[width=0.17\textwidth]{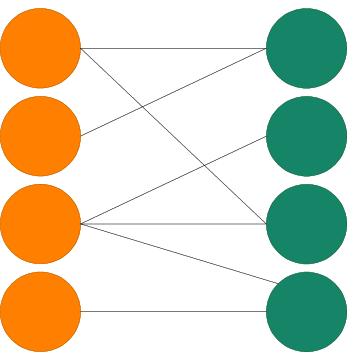}\\
         & & & $BP$ -- bipartite \\
         & & & \includegraphics[width=0.17\textwidth]{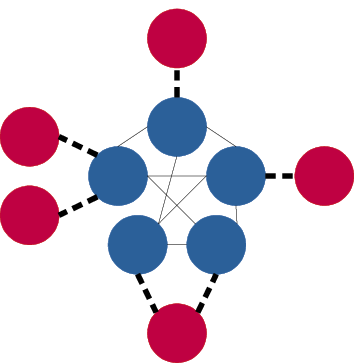} \\
        $G_3$ & $G_6$ & $C2$ & $BP$ -- hub
    \end{tabular}
    \caption{Illustrative examples of the synthetic graphs used. Solid (resp. dashed) lines indicate intra (resp. inter)-community edges. Vertices of the same color belong to the same ground truth community. Examples given for $G_3$, $C2$ and $BP$ -- bipartite have a cluster size of 4 vertices; those given for $G_6$ and $BP$ -- hub have a size of 5 vertices. Cluster sizes were chosen to be small for the purpose of visualization in this figure.}
    \label{fig:synth_examples}
\end{figure}}

\paragraph{Generating observation and model graphs:} for each synthetic graph, the observation graph $G_O$ was first built by creating the specified amount of vertices per community. Then, for each pair of vertices, an edge may be added or not, with the probability of the edge being added specified by the communities to which each vertex in the pair belongs. 

The model graph $G_M$ was generated by adding the expected value of connections between communities -- thus, the diagonal values (that is, self-loops) are ${a_M}_{jj} = 2R_{jj}\cdot s_j$ (thus encoding the amount of intra-connections in $j$) and the non-diagonal values are ${a_M}_{jk} = R_{jk}\cdot (s_j + s_k)$ (thus encoding the amount of inter-connections between $j$ and $k$), where $R_{jk}$ is the probability of connecting communities $j$ and $k$ and $s_j$ is the size of community $j$.

\subsubsection{Experimental Setup.}
\label{sssec:synth_experimental_setup}

Three main experiments were set up:
\paragraph{Basic Graphs Comparison:} A preliminary experiment was performed on the $G_3$, $G_6$ and $C2$ datasets, to compare the performance of each method on basic scenarios. The central inter-community probability for $C2$ was set to $0.42$. Different sizes for the communities were explored: $\{5, 10, 20, 40, 80\}$. We aimed to analyze the basic behavior of all techniques on graphs of different sizes.

\begin{figure}[htbp]
    \centering
    \begin{tabular}{ccccc}
        \includegraphics[width=0.19\textwidth]{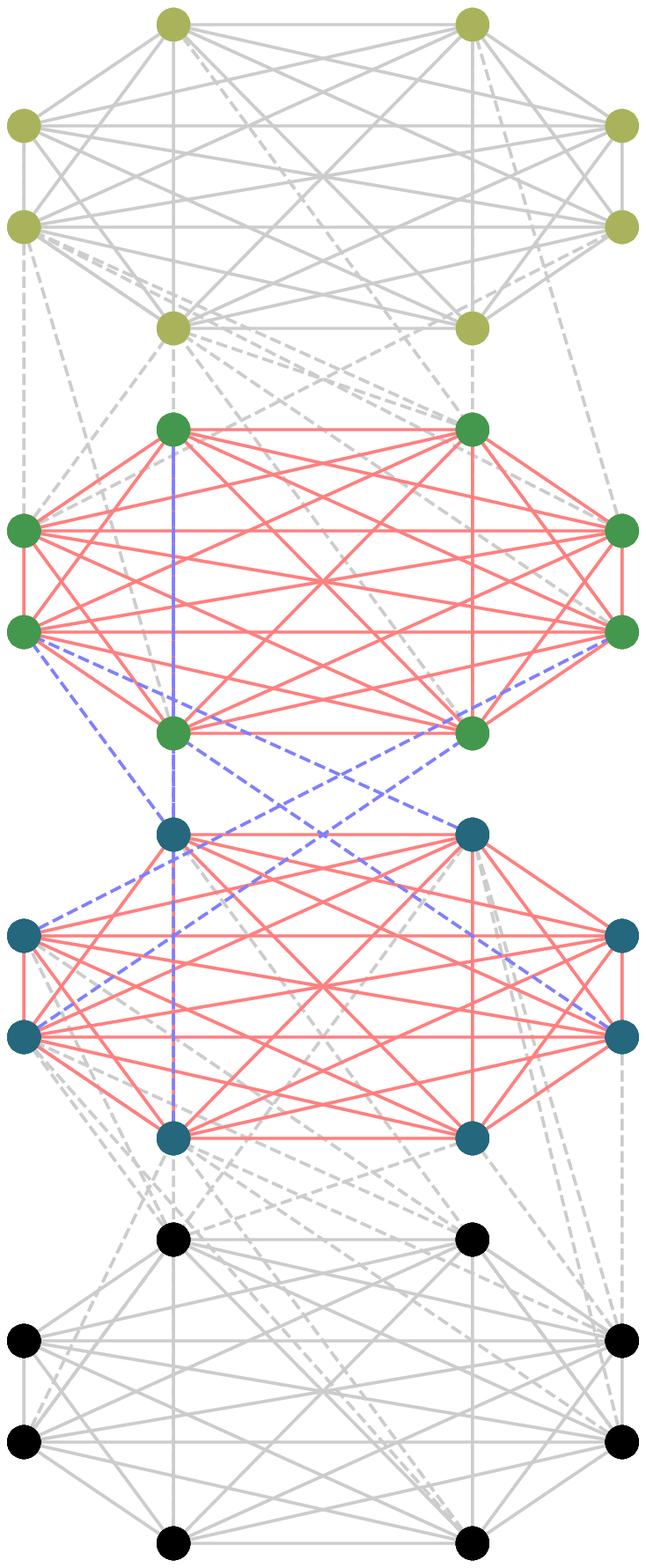} & 
        \includegraphics[width=0.19\textwidth]{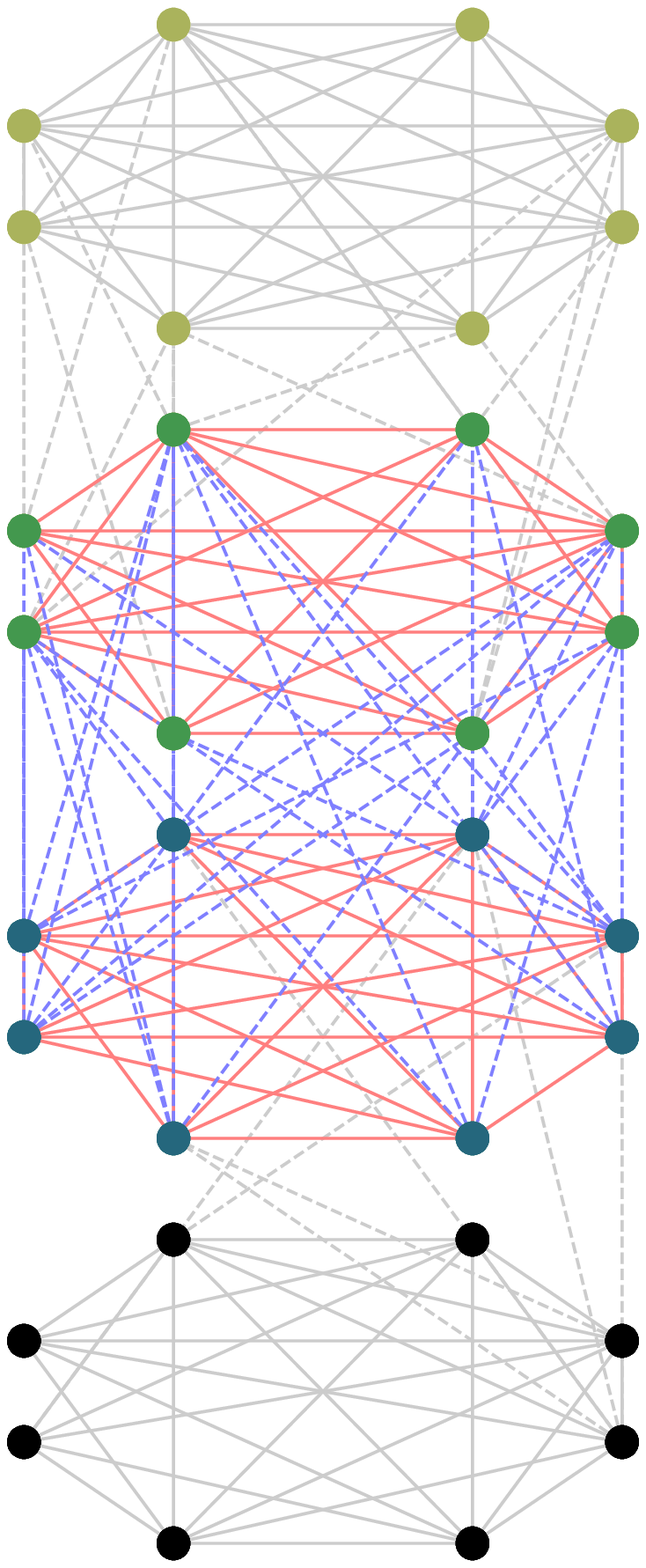} &
        \includegraphics[width=0.19\textwidth]{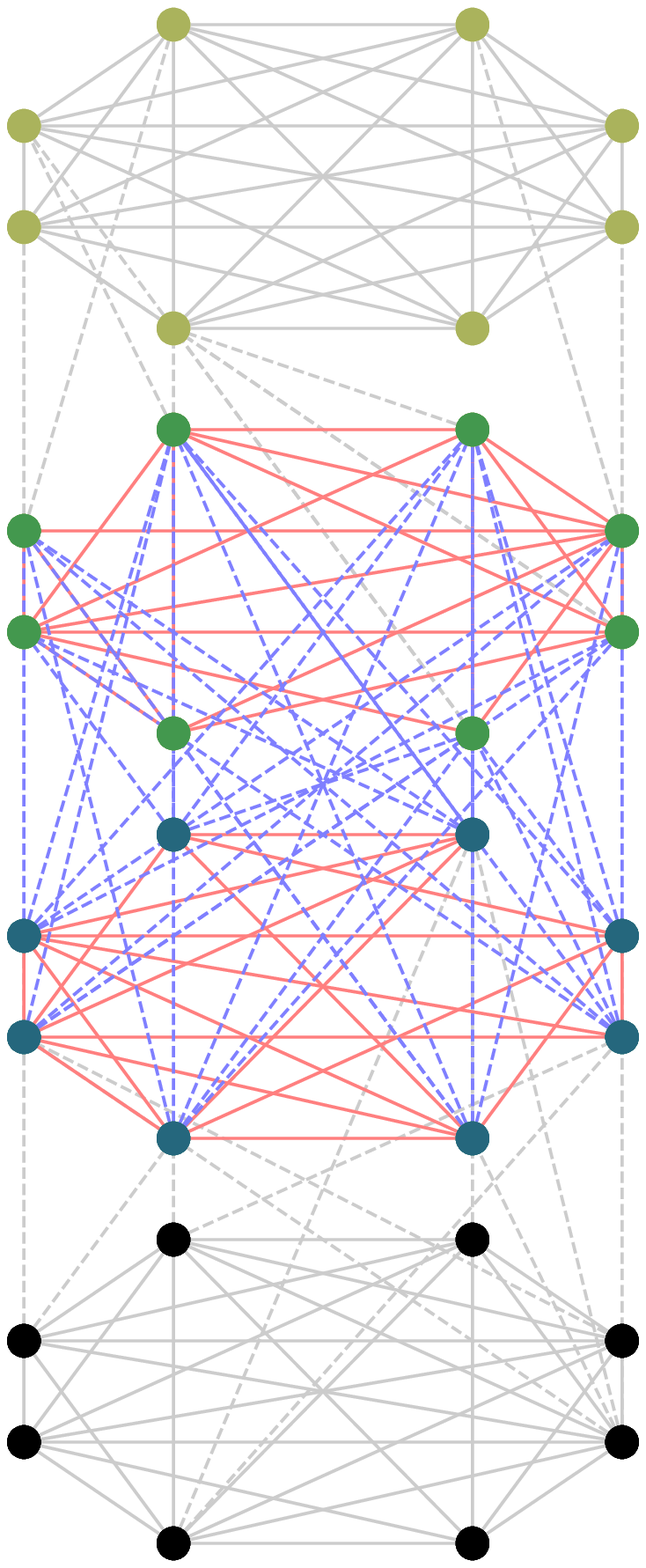} &
        \includegraphics[width=0.19\textwidth]{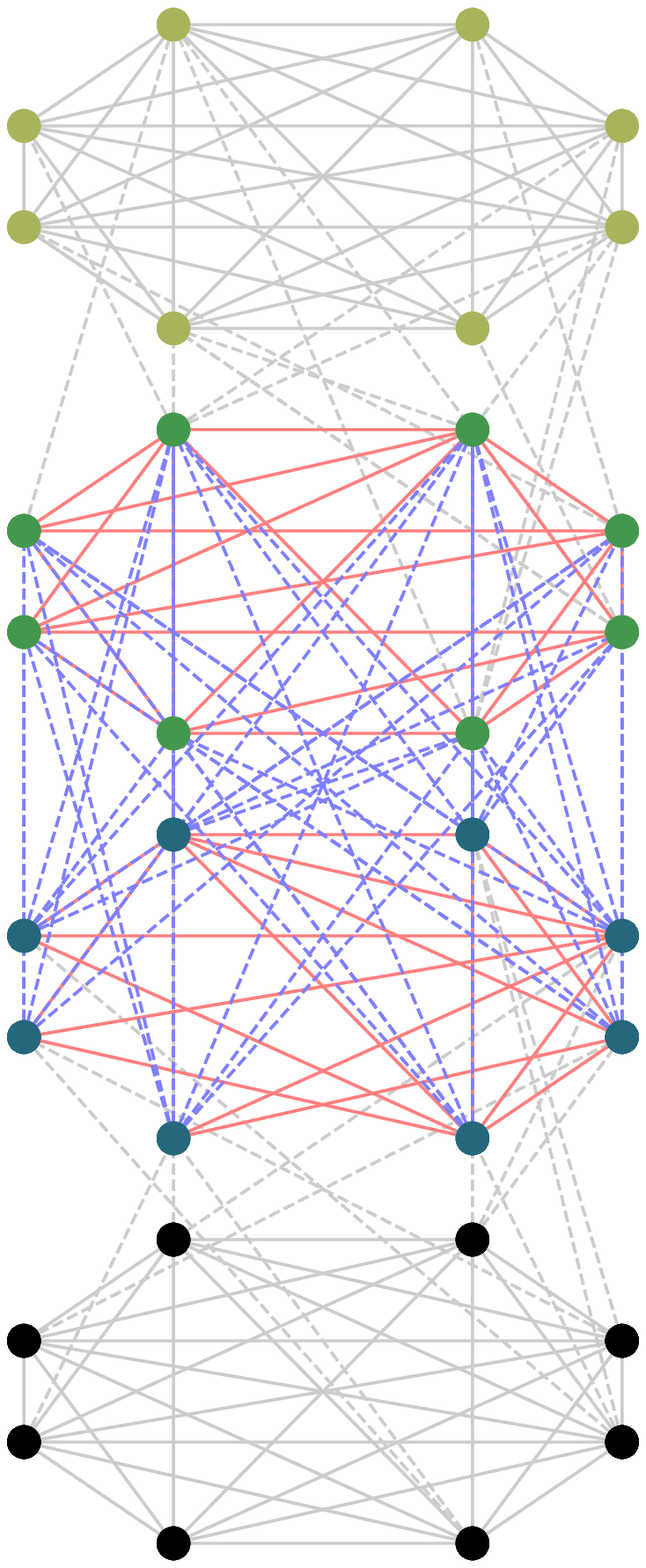} &
        \includegraphics[width=0.19\textwidth]{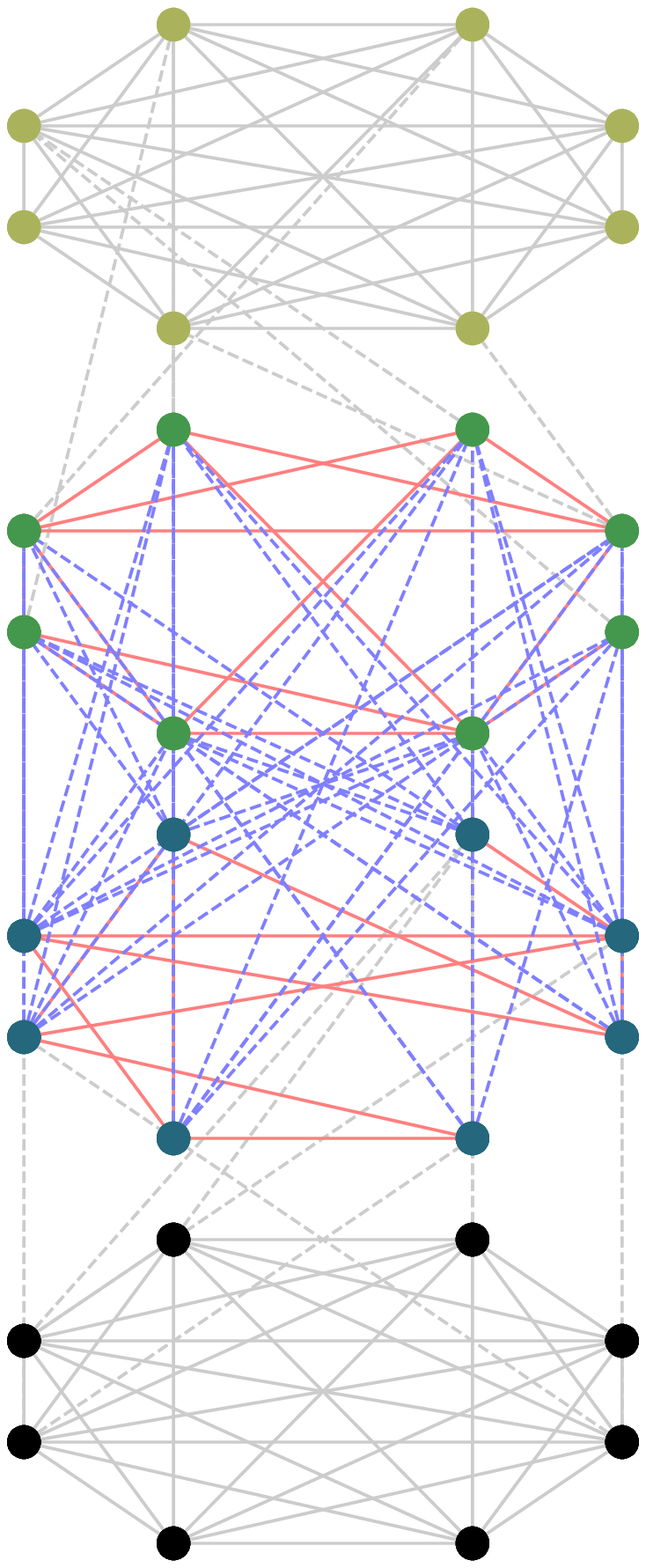} \\
       \vtop{\hbox{\strut 0.8 intra}\hbox{\strut 0.1 inter}} & 
       \vtop{\hbox{\strut 0.6 intra}\hbox{\strut 0.3 inter}} & 
       \vtop{\hbox{\strut 0.5 intra}\hbox{\strut 0.4 inter}} & 
       \vtop{\hbox{\strut 0.4 intra}\hbox{\strut 0.5 inter}} & 
       \vtop{\hbox{\strut 0.3 intra}\hbox{\strut 0.6 inter}}
    \end{tabular}
    \caption{Examples of progressively harder-to-separate $C2$ graphs with cluster size of 8. Vertices of the same color belong to the same community. Solid (resp. dashed) lines indicate intra (resp. inter)-community edges. Gray lines indicate edges with an endpoint in the ``outside'' communities. Red lines are intra-community edges within the central communities; blue lines are inter-community edges connecting the central communities. Each subcaption shows the probability of inter- and intra-connections in the two central communities. From left to right, as the probabilities shift towards harder-to-separate scenarios, we see less red edges and more blue edges.\label{fig:C2_progression_example}}
\end{figure}

\paragraph{$C2$ Progressive Difficulty:} Experiments were performed on the $C2$ graph exploring the following values for the inter-connection probability of the central communities: $\{.20,.25,.30,.35,.40,.45,.50,.55,.60\}$. Different sizes for the communities were explored: $\{10, 20, 40\}$. The intent of this experiment is to analyze the behavior of all clustering techniques when faced with progressively harder-to-separate scenarios, i.e. when inter-connection probabilities increase, the two central communities become more and more indistinguishable. A visual example of progressively harder scenarios is shown in Figure~\ref{fig:C2_progression_example}. 

\begin{figure}[htbp]
    \centering
    \begin{tabular}{ccc}
        \includegraphics[width=0.2\textwidth,trim={20 30 20 40},clip]{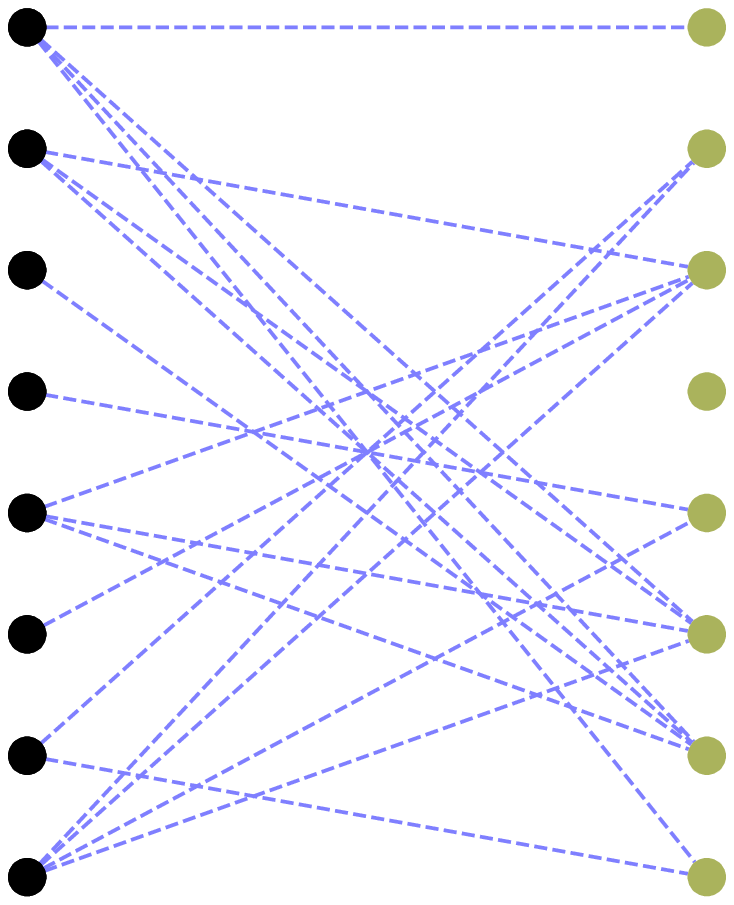} &
        \includegraphics[width=0.2\textwidth,trim={20 30 20 40},clip]{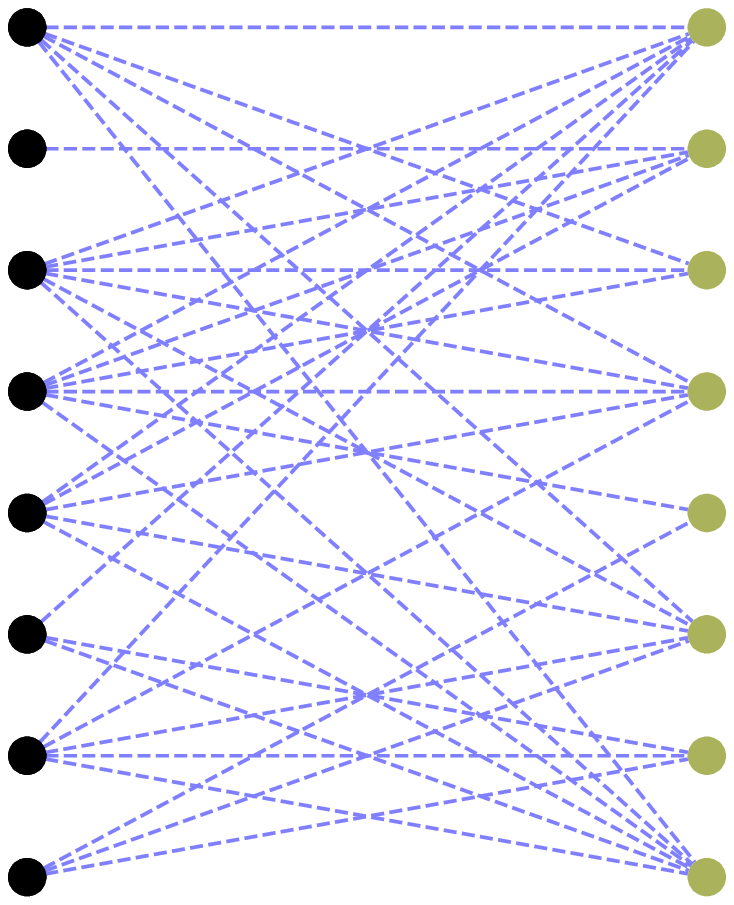} &
        \includegraphics[width=0.2\textwidth,trim={20 30 20 40},clip]{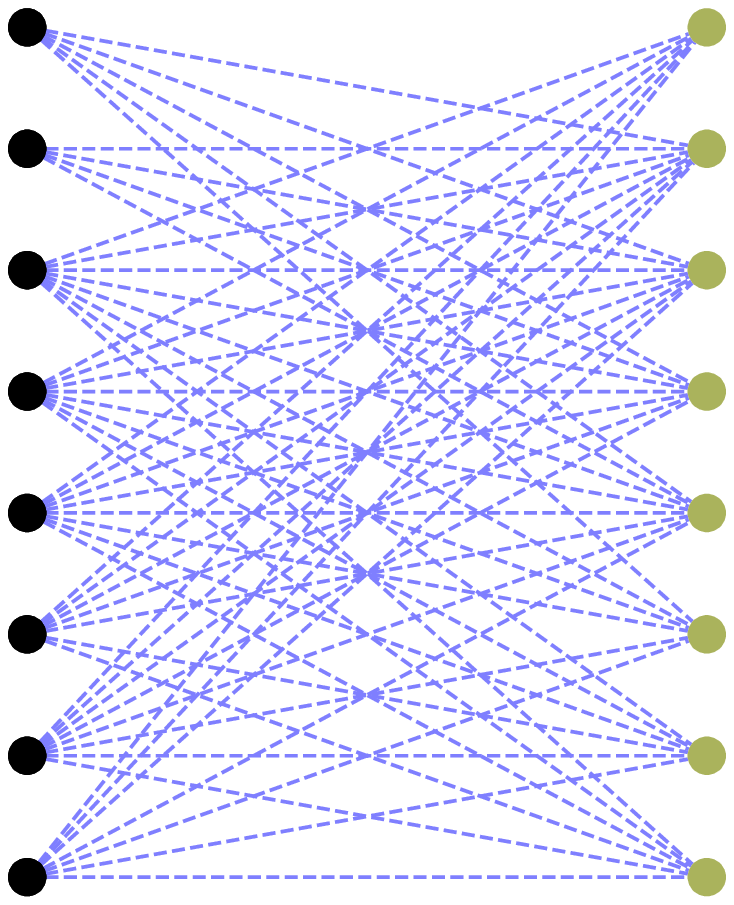} \\
        \includegraphics[width=0.25\textwidth,trim={15 15 15 40},clip]{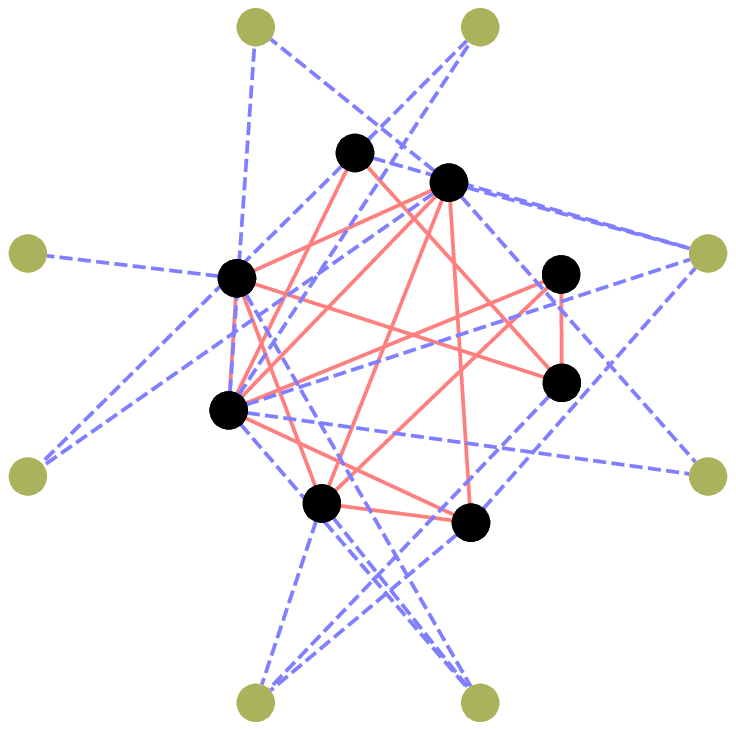} &
        \includegraphics[width=0.25\textwidth,trim={15 15 15 40},clip]{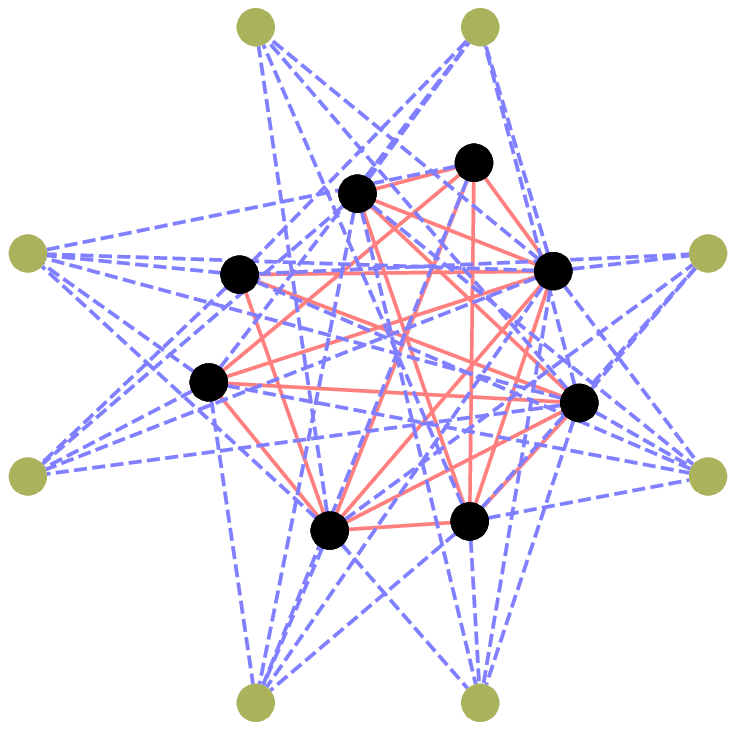} &
        \includegraphics[width=0.25\textwidth,trim={15 15 15 40},clip]{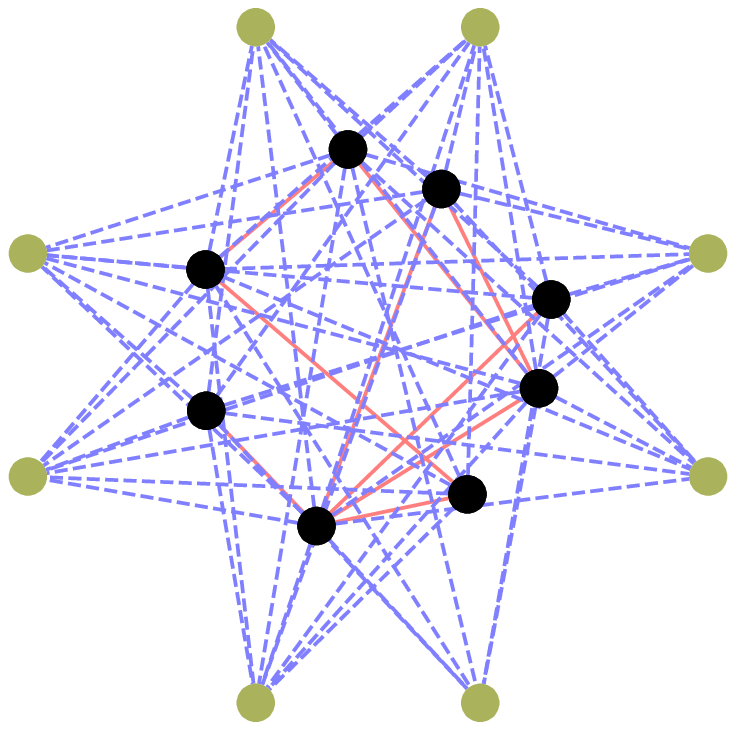} \\
        0.3 inter & 0.6 inter & 0.9 inter  \\ 
    \end{tabular}
    \caption{Examples of progressively harder-to-separate $BP$ -- bipartite (top row) and $BP$~-- hub (bottom row) graphs with cluster size of 8. Vertices of the same color belong to the same community. Solid (resp. dashed) lines indicate intra (resp. inter)-community edges. Red lines are intra-community edges; blue lines are inter-community edges. Each subcaption shows the probability of inter-connections between communities. From left to right, as the probabilities increase, we see more blue edges.}
    \label{fig:BP_examples}
\end{figure}

\paragraph{Bipartite and Hub Comparison:} Experiments were performed on the $BP$ graph with values $\{.40,.45,.50, .55,.60,.65,.70,.75,.80\}$ for the inter-connection probabilities, and  two values for the intra-connection probabilities: $0$ (or ``Bipartite'') and $.50$ (or ``Hub''). Different sizes for the communities were explored: $\{10, 20, 40\}$. The intent of this experiment was to analyze the behavior of our clustering technique in a scenario that is known to be extremely hard for traditional spectral clustering to perform. A visual example of both hubs and bipartite graphs with different inter-connection probabilities can be seen in Figure~\ref{fig:BP_examples}.

\subsubsection{Results and Discussion.}
\label{sssec:synth_results}
\begin{figure}[htbp]
    \centering
    \begin{tabular}{cc}
         \includegraphics[width=0.5\textwidth]{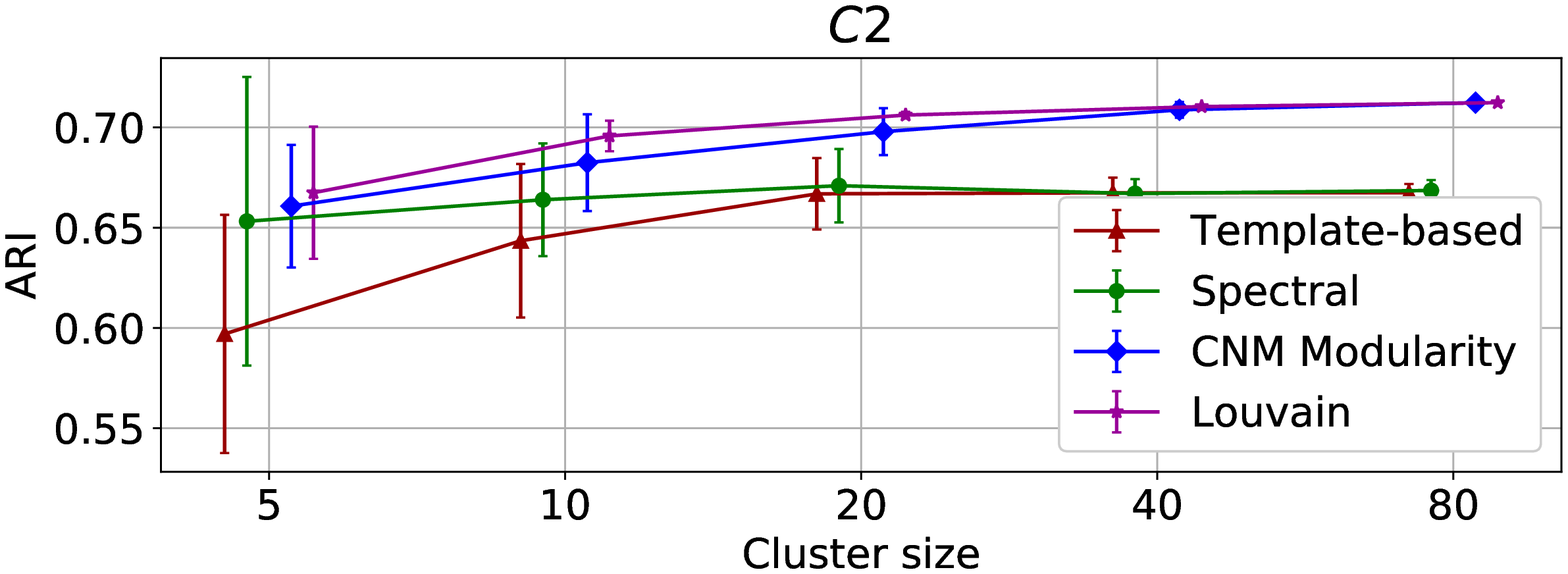} &
         \includegraphics[width=0.5\textwidth]{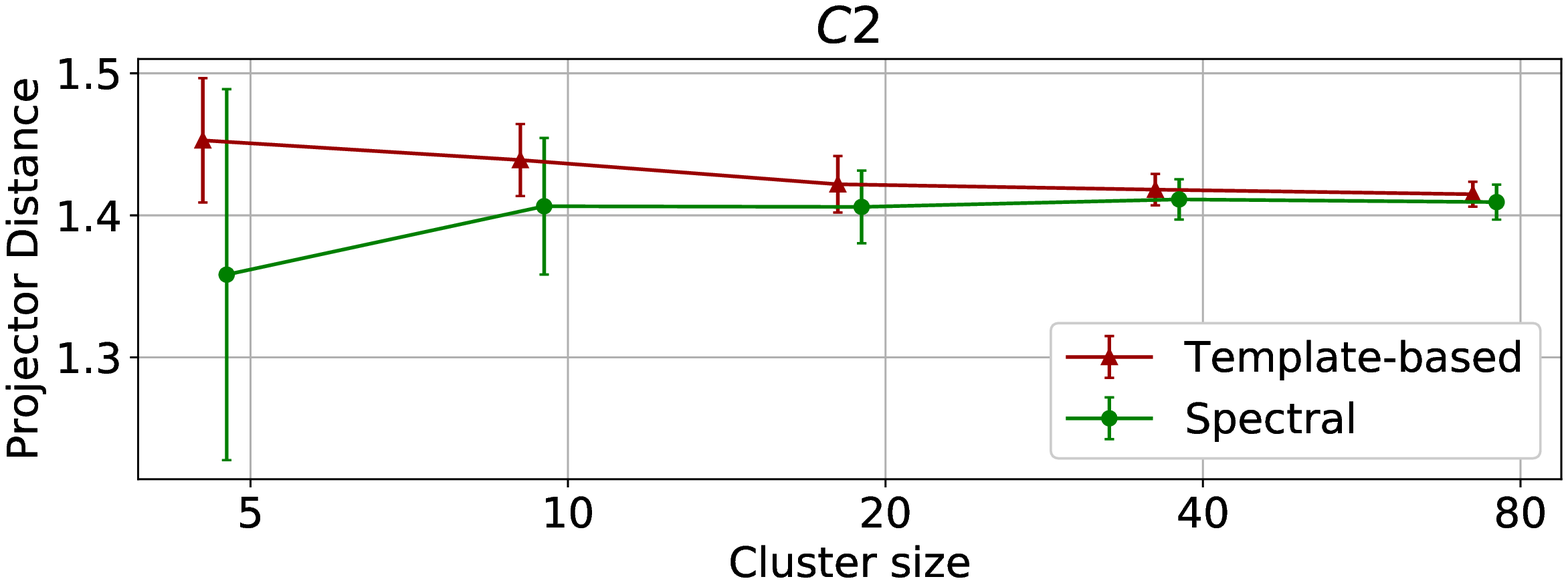} \\
         \includegraphics[width=0.5\textwidth]{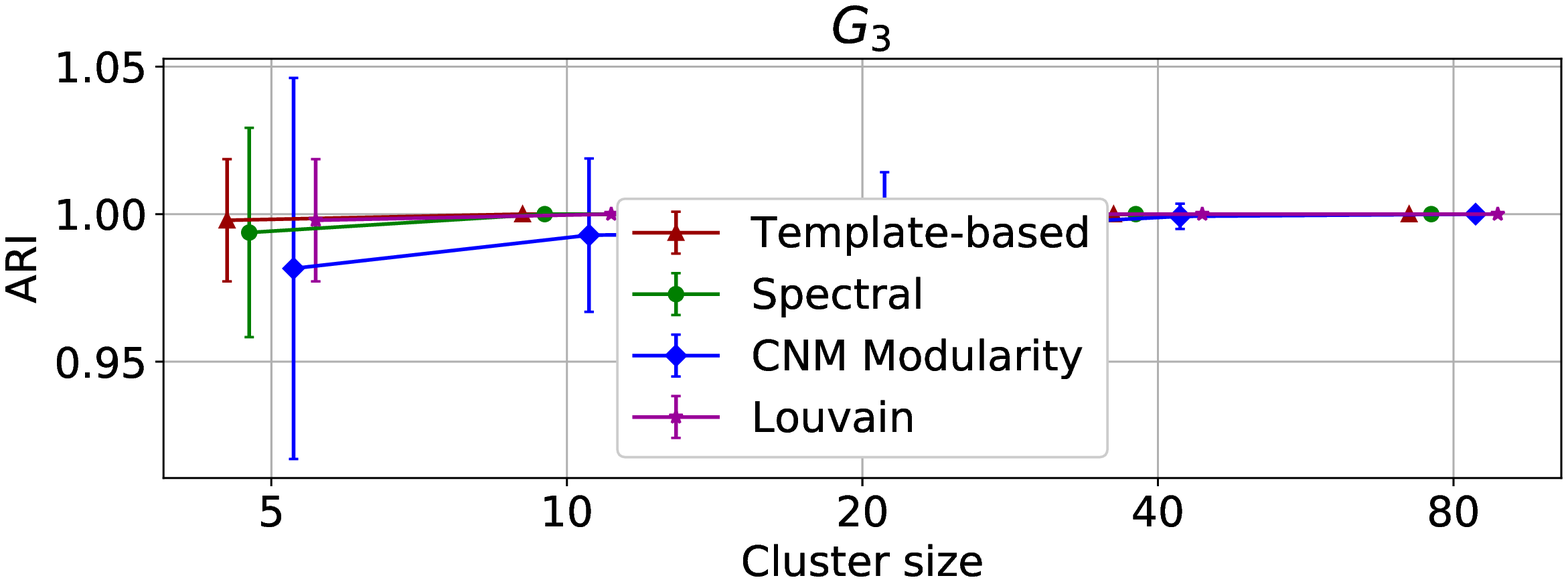} &
         \includegraphics[width=0.5\textwidth]{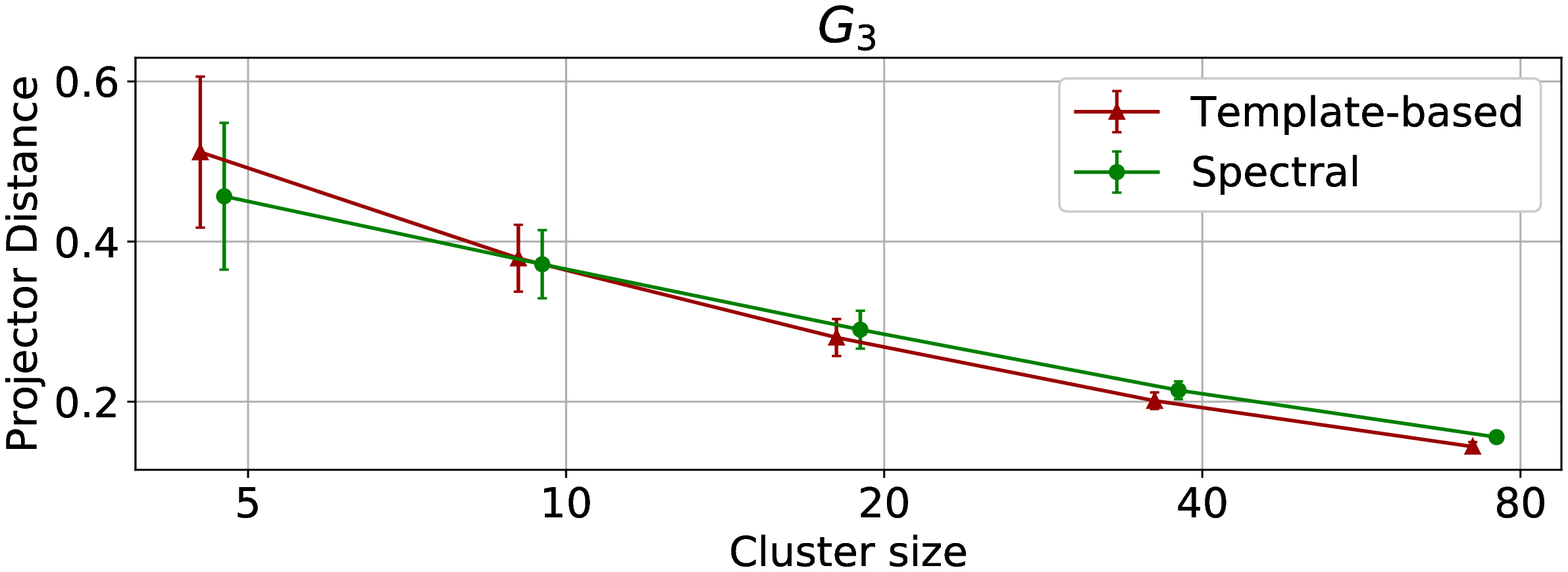} \\
         \includegraphics[width=0.5\textwidth]{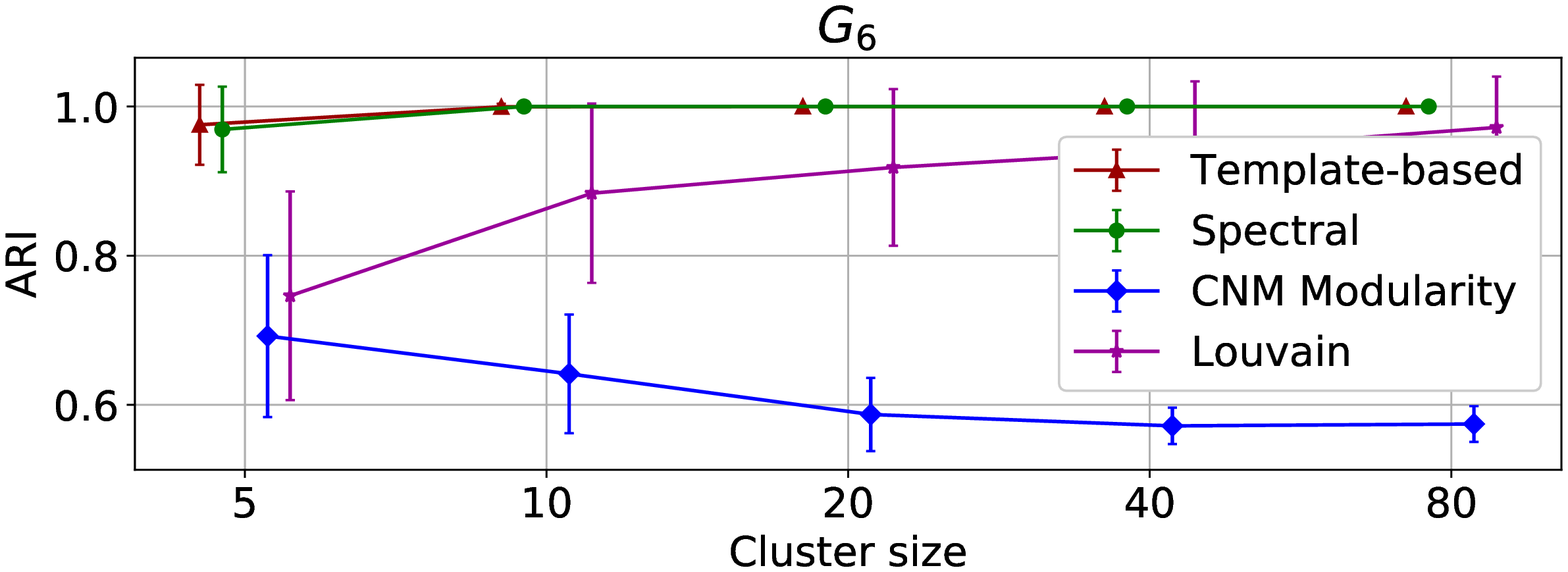}  &
         \includegraphics[width=0.5\textwidth]{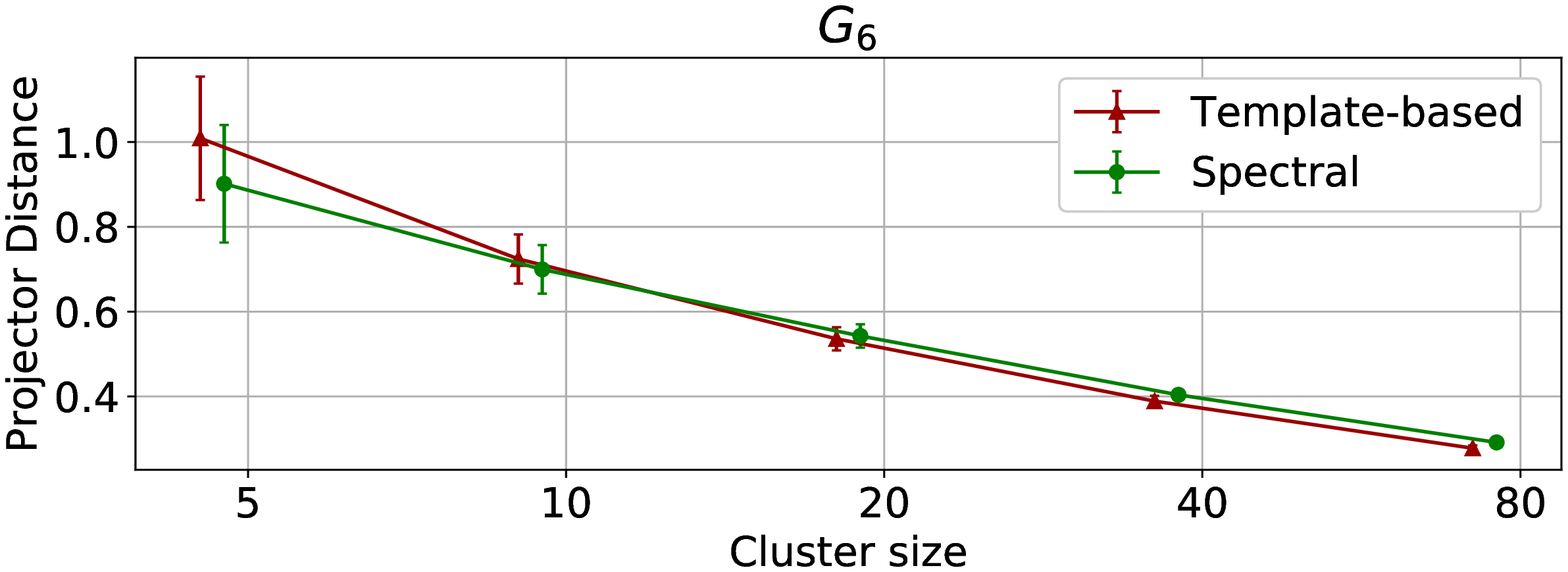} 
    \end{tabular}
    \caption{Adjusted Rand Index and Projector Distance for the Basic Graphs Comparison experiments. Error bars represent the standard deviation of the results. Rows correspond to different graphs ($C2$, $G_3$, and $G_6$, respectively).}
    \label{fig:basic_results}
\end{figure}

\begin{figure}[htbp]
    \centering
    \begin{tabular}{ccc}
        \includegraphics[width=0.3\textwidth]{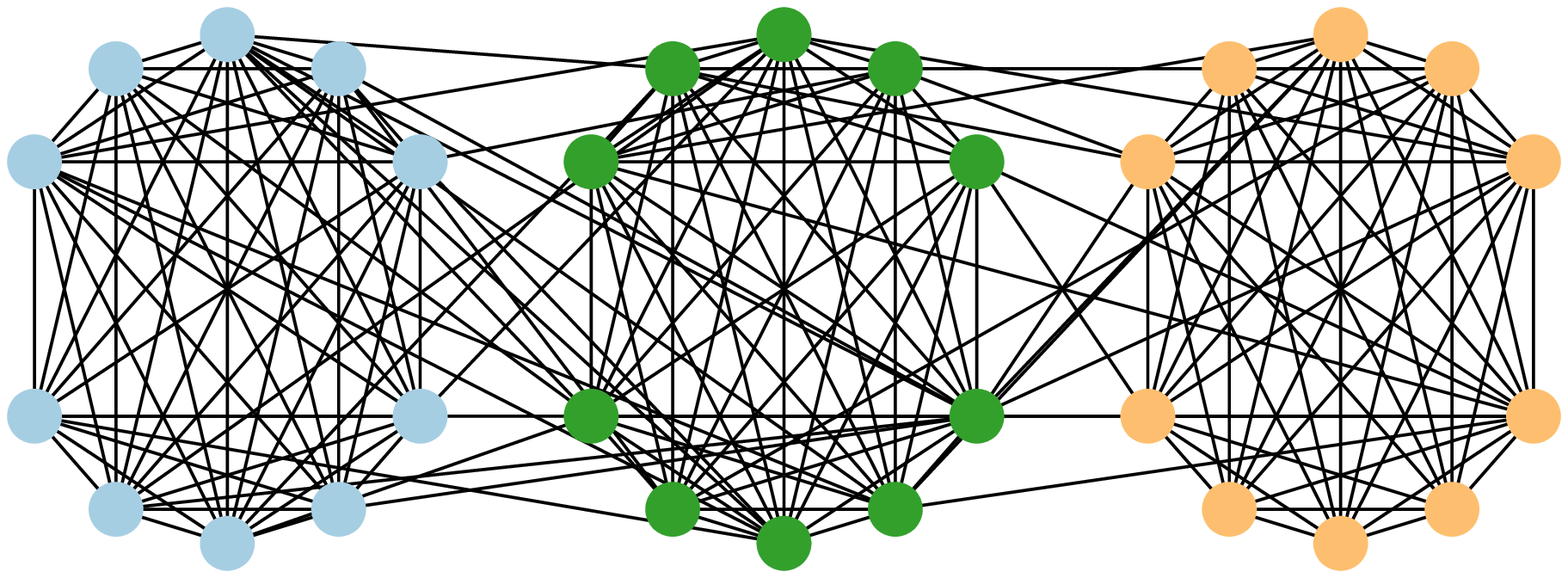} & \includegraphics[width=0.3\textwidth]{figures/qualitative_G3.eps} &
        \includegraphics[width=0.3\textwidth]{figures/qualitative_G3.eps} \\
        $G_3$ for TB: ARI $1.0$ & $G_3$ for spectral: ARI $1.0$ & $G_3$ for modularity: ARI $1.0$ \\
         \includegraphics[width=0.3\textwidth]{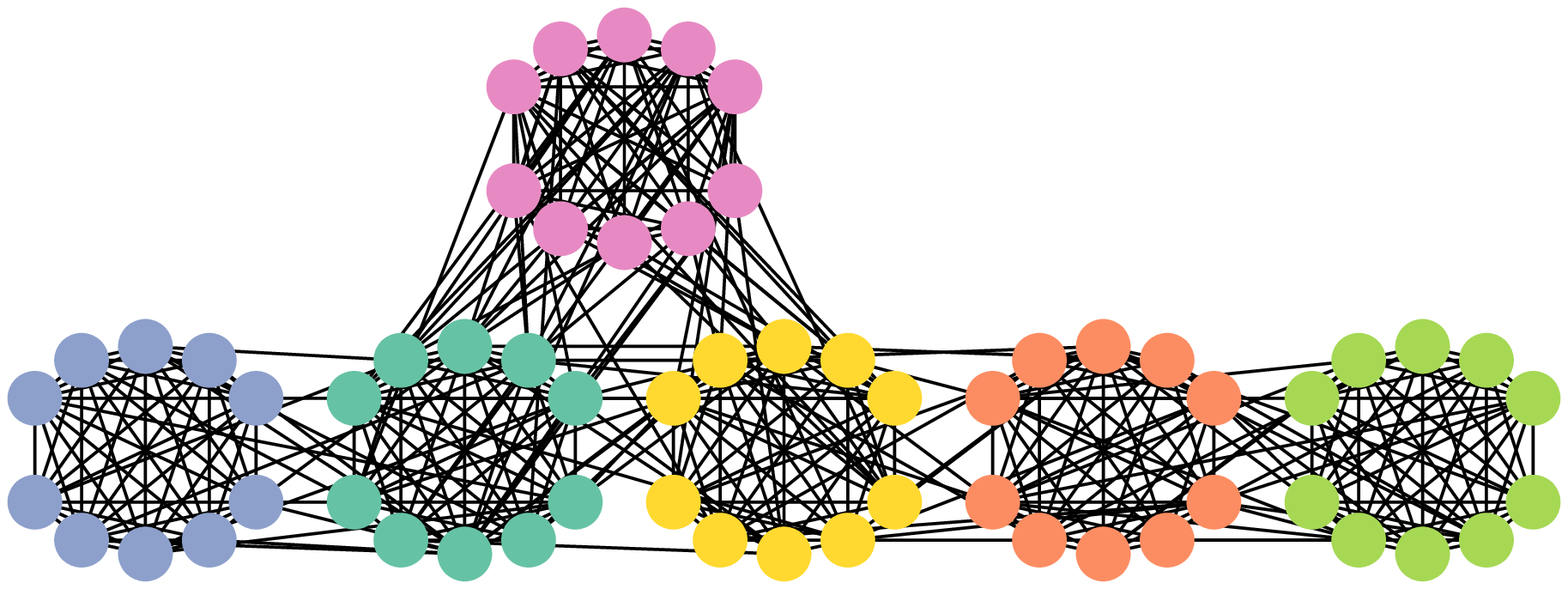} &
         \includegraphics[width=0.3\textwidth]{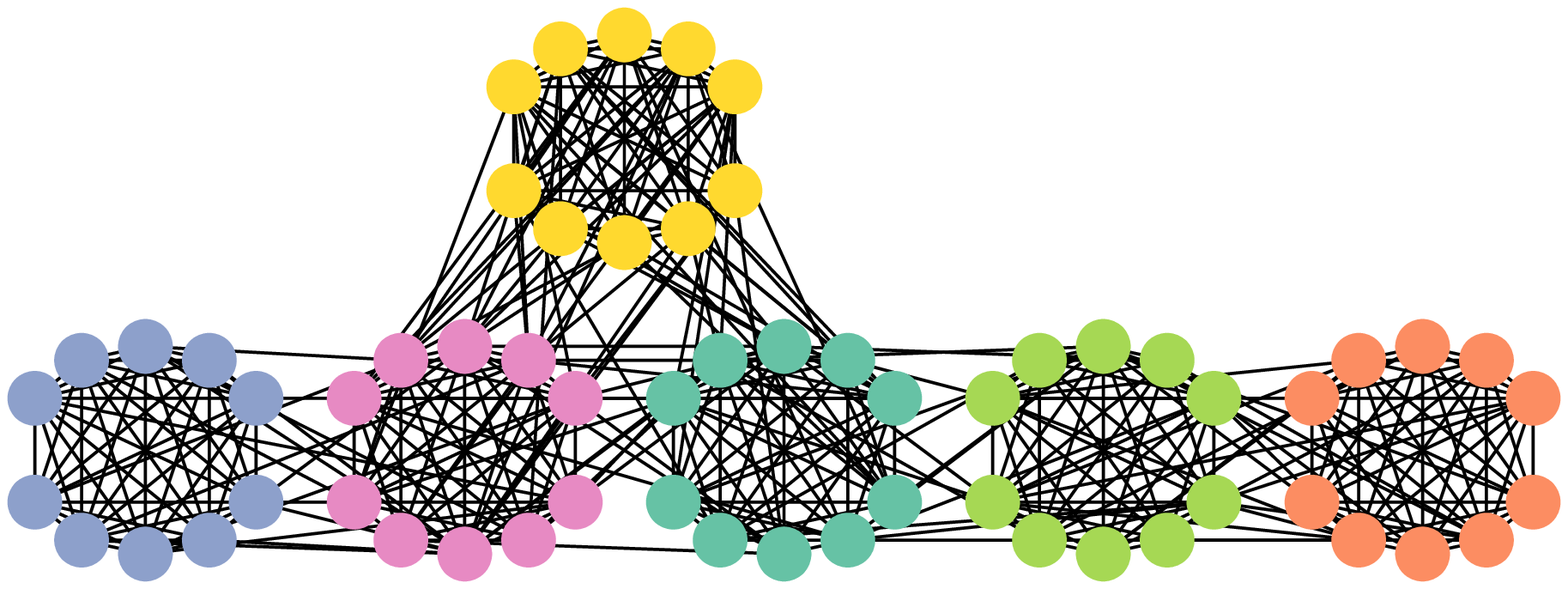} &
         \includegraphics[width=0.3\textwidth]{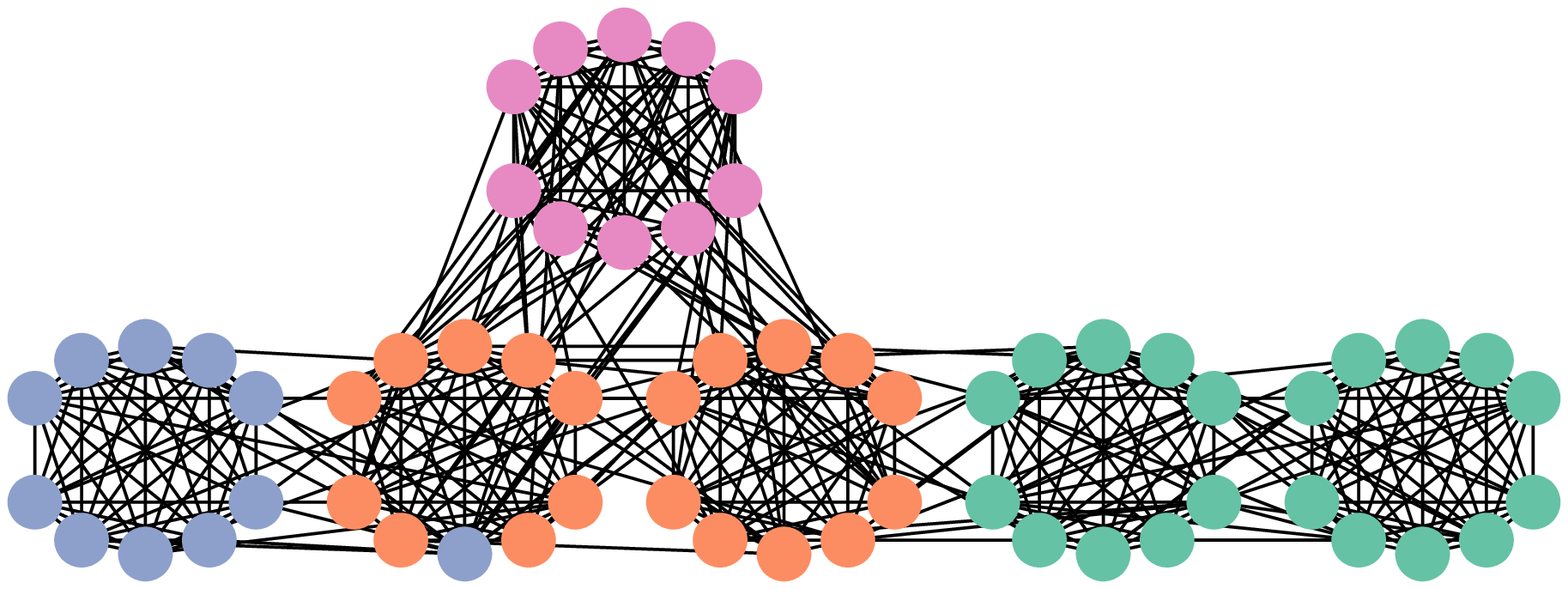} \\
         $G_6$ for TB: ARI $1.0$ & $G_6$ for spectral: ARI $1.0$ & $G_6$ for modularity: ARI $0.64$ \\
         \includegraphics[width=0.3\textwidth]{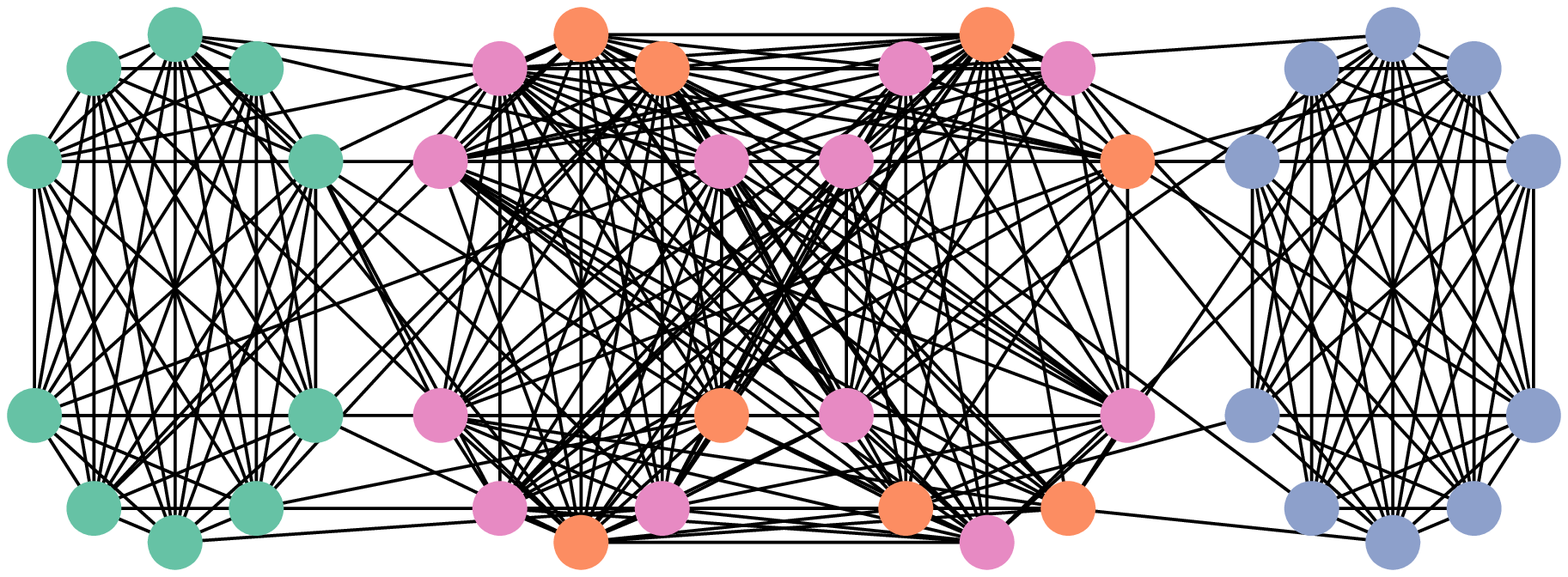} &
         \includegraphics[width=0.3\textwidth]{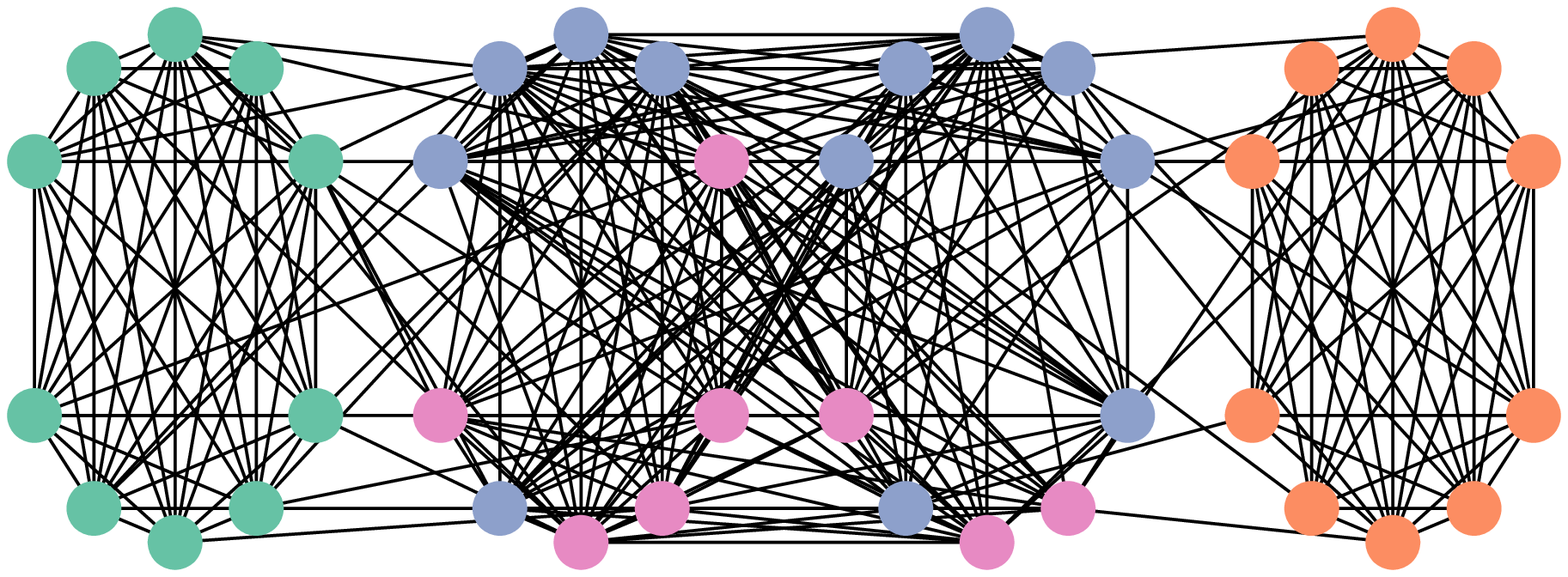} &
         \includegraphics[width=0.3\textwidth]{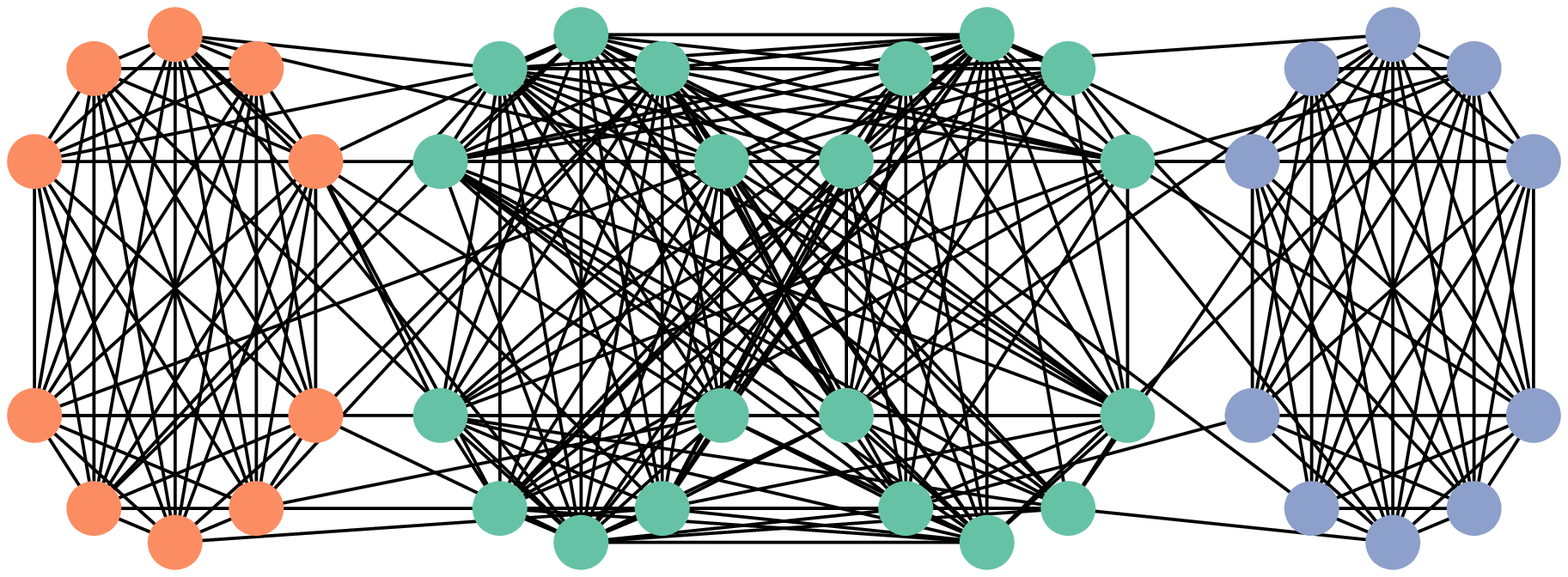} \\
         $C2$ for TB: ARI $0.64$ & $C2$ for spectral: ARI $0.66$ & $C2$ for modularity: ARI $0.7$ \\
    \end{tabular}
    \caption{Qualitative results for the basic experiment with cluster size 10. Each circle of vertices is a ground truth community; vertices of the same color are predicted to be in the same community. 
    }
    \label{fig:basic_qualitative}
\end{figure}

\emph{Basic Graphs Comparison Results:} Figure~\ref{fig:basic_results} shows the results for the Basic Graphs Comparison experiment. Figure~\ref{fig:basic_qualitative} displays qualitative results for each method and graph.
The ARI of the TB clustering is similar to the spectral clustering (baseline) in most scenarios, with notable exceptions of small-clusters $C2$, where it under-performs. The projector distance of the TB method diminishes in all cases and decreases faster than the spectral method as the size of clusters increases, outperforming spectral clustering in $G_3$ and $G_6$ starting at cluster size 20.
Modularity performance is highly dependent on the graph being segmented: for $C2$, the performance is increased as the method segments external communities correctly and the central communities together, as can be seen in the qualitative result. For $G_6$, however, the performance for CNM is significantly lower, mainly because modularity does not force the existence of all clusters. Louvain avoids this shortcoming, but still underperforms until cluster size increases.

\begin{figure}[htbp]
    \centering
    \begin{tabular}{cc}
         \includegraphics[width=0.5\textwidth]{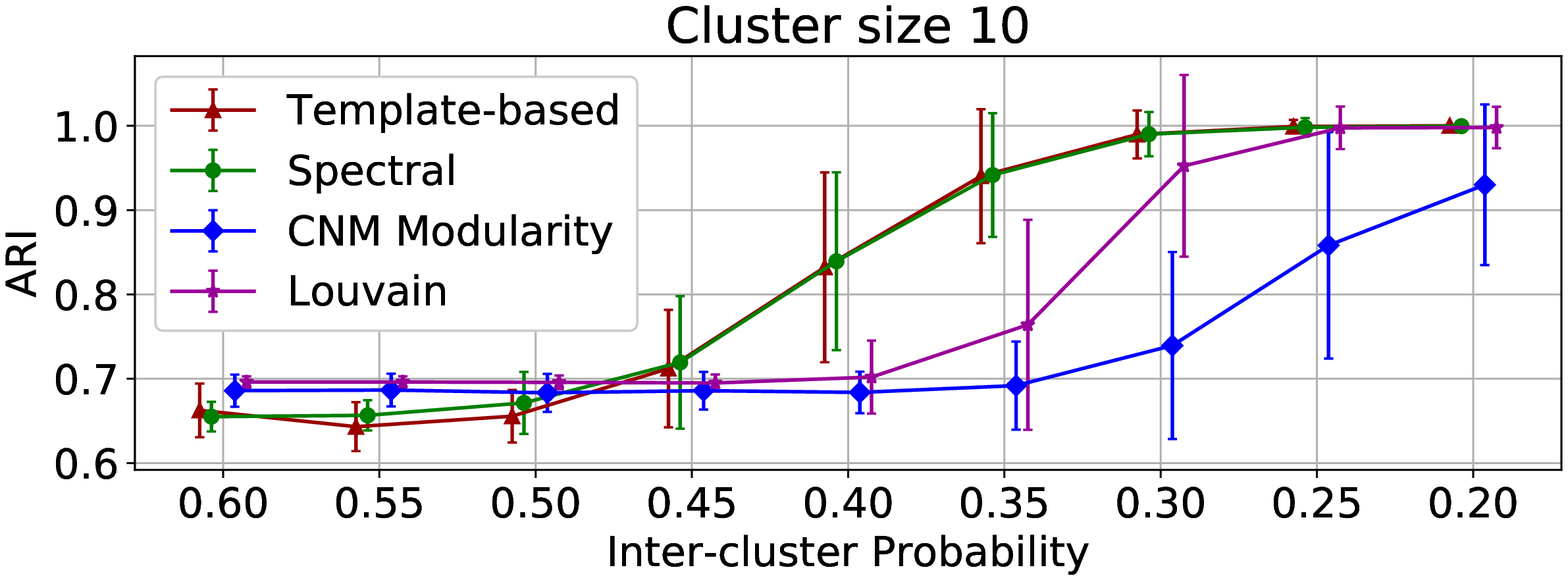} &
         \includegraphics[width=0.5\textwidth]{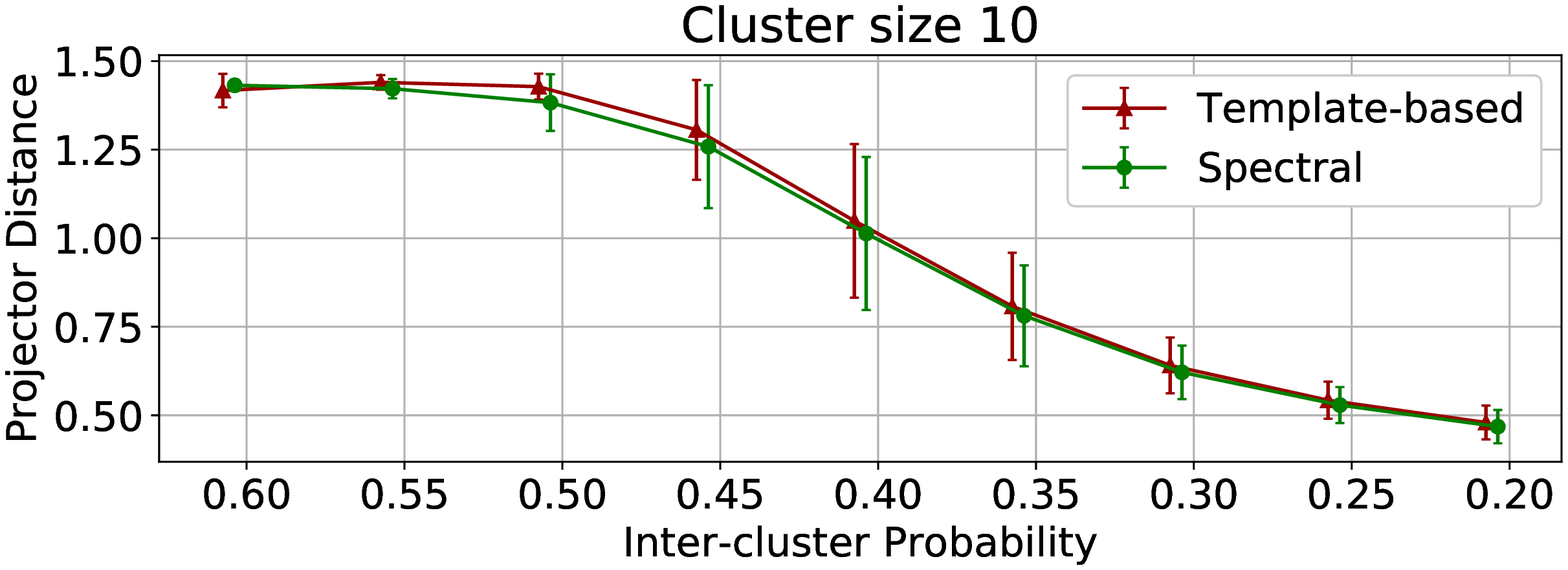} \\
         \includegraphics[width=0.5\textwidth]{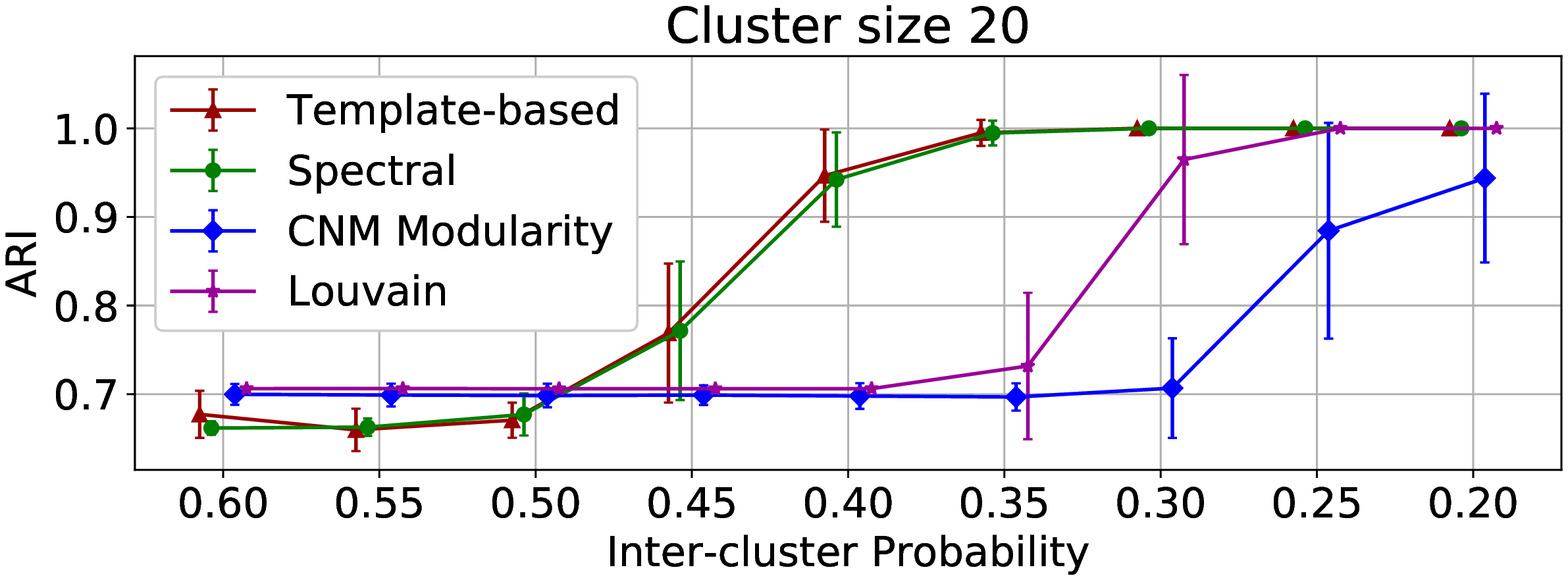} &
         \includegraphics[width=0.5\textwidth]{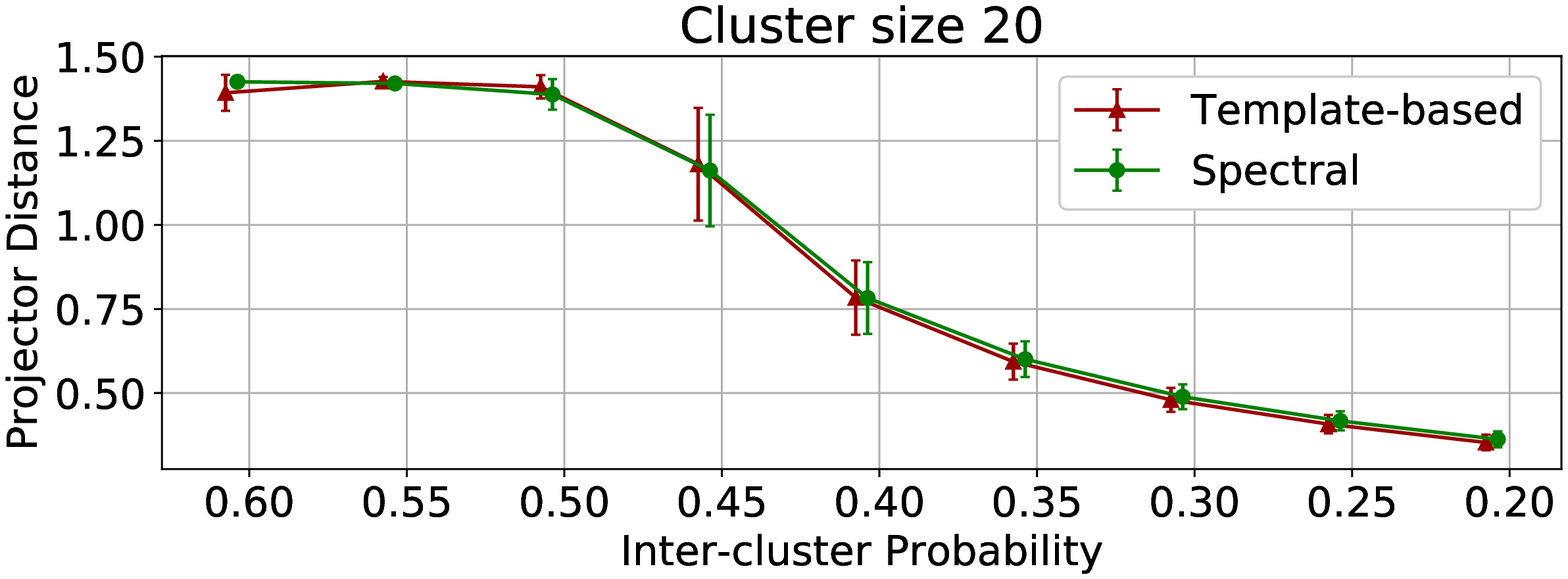} \\
         \includegraphics[width=0.5\textwidth]{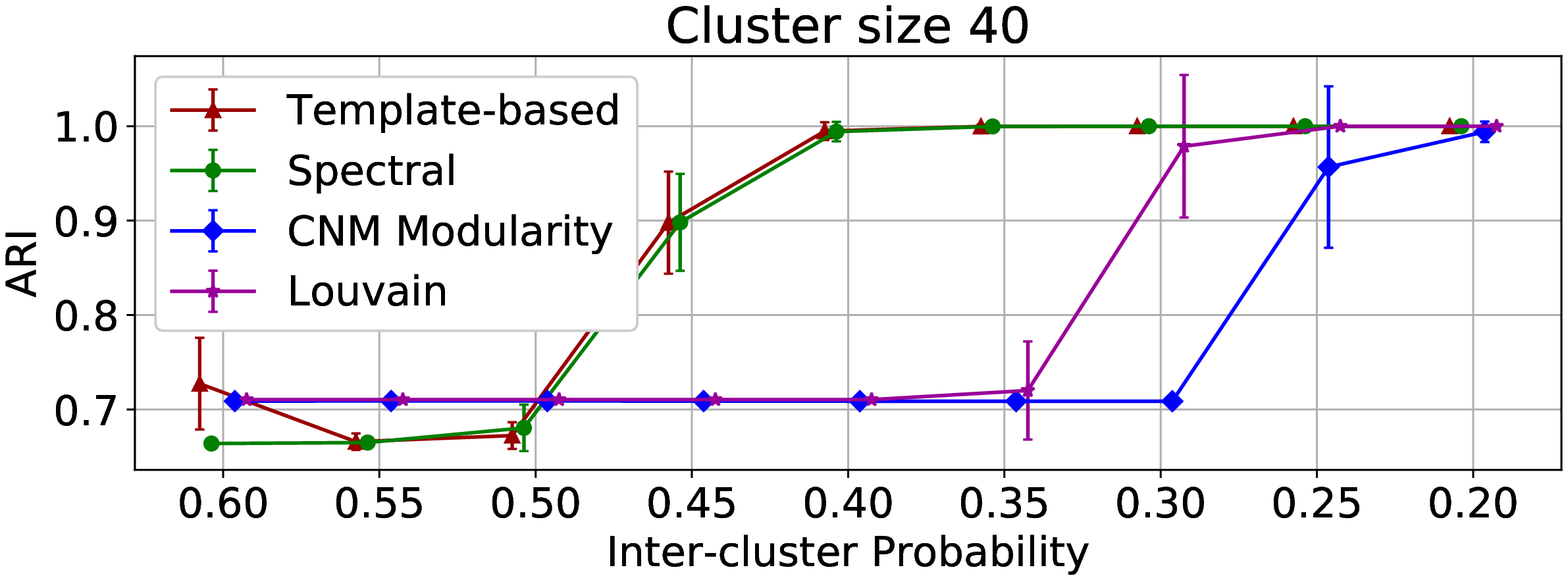} &
         \includegraphics[width=0.5\textwidth]{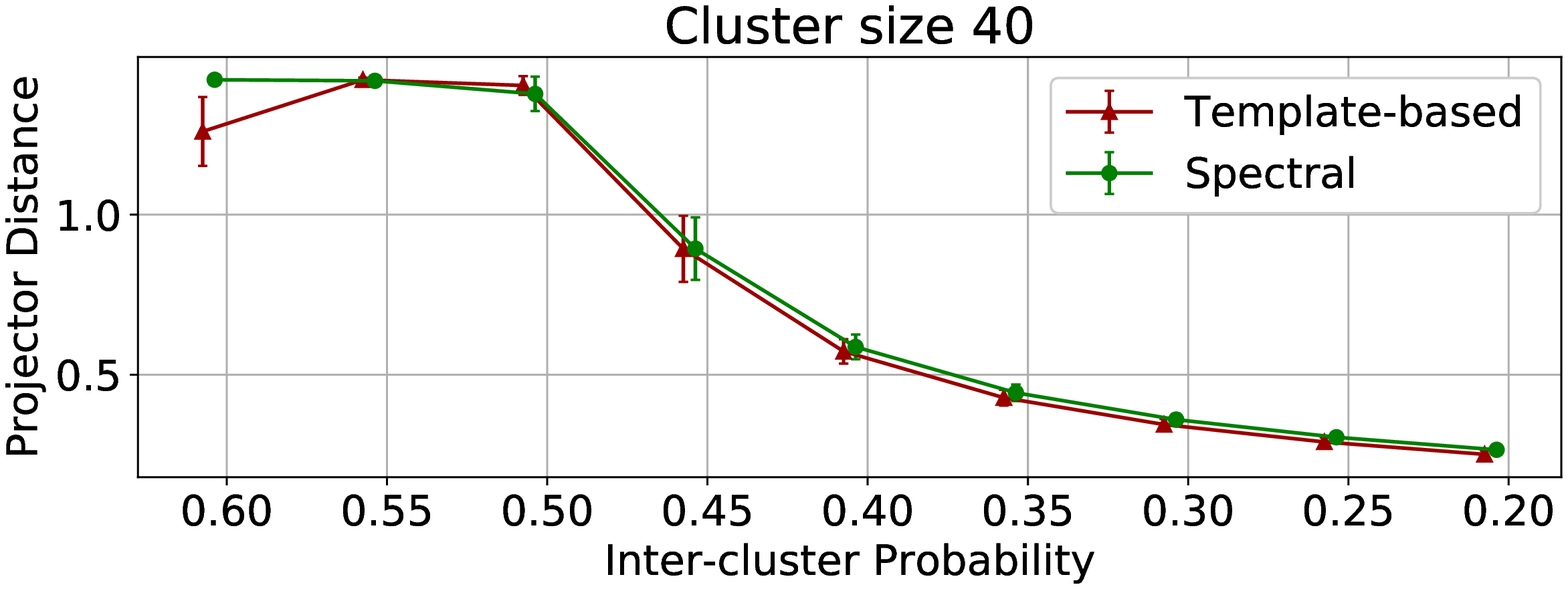}
    \end{tabular}
    \caption{Adjusted Rand Index and Projector Distance for the $C2$ Progressive Difficulty experiments. Error bars represent the standard deviation of the results. Rows are for different cluster sizes (10, 20 and 40).}
    \label{fig:progression_results}
\end{figure}

\emph{C2 Progressive Difficulty Results:} Figure~\ref{fig:progression_results} shows the results for the C2 Progressive Difficulty experiment.
We can note, again, that both TB and spectral graph clustering perform similarly, except on the hardest case (when inter-cluster probability is $.60$) with results being better the greater the cluster size. Modularity, again, under-performs when the difficulty is low before stabilizing at predicting three clusters, joining the central clusters.

\begin{figure}[htbp]
    \centering
    \begin{tabular}{cc}
         \includegraphics[width=0.5\textwidth]{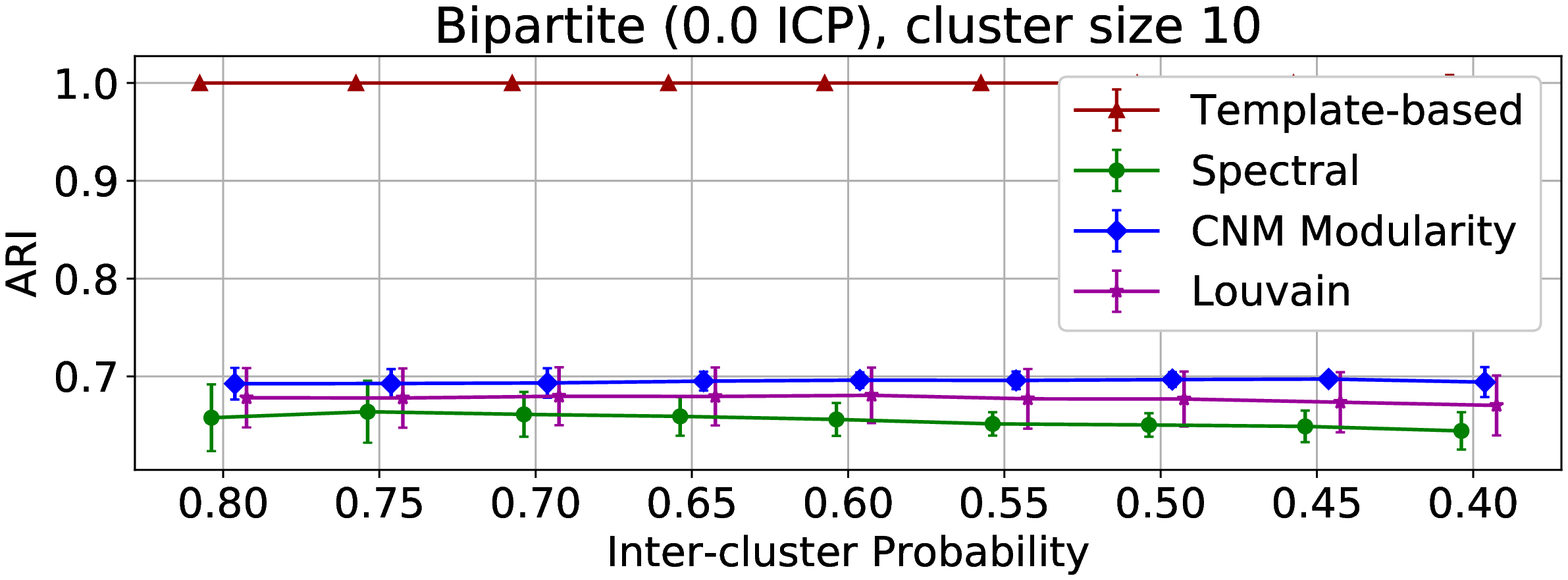} &
         \includegraphics[width=0.5\textwidth]{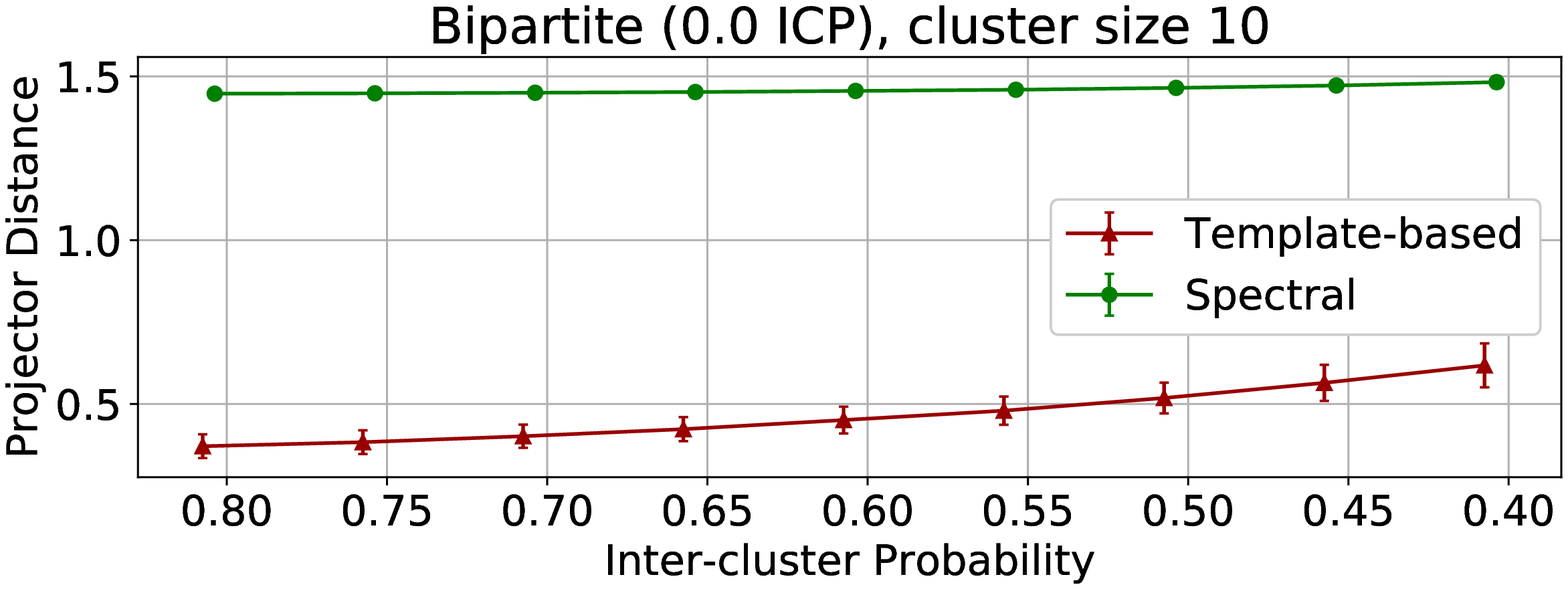} \\
         \includegraphics[width=0.5\textwidth]{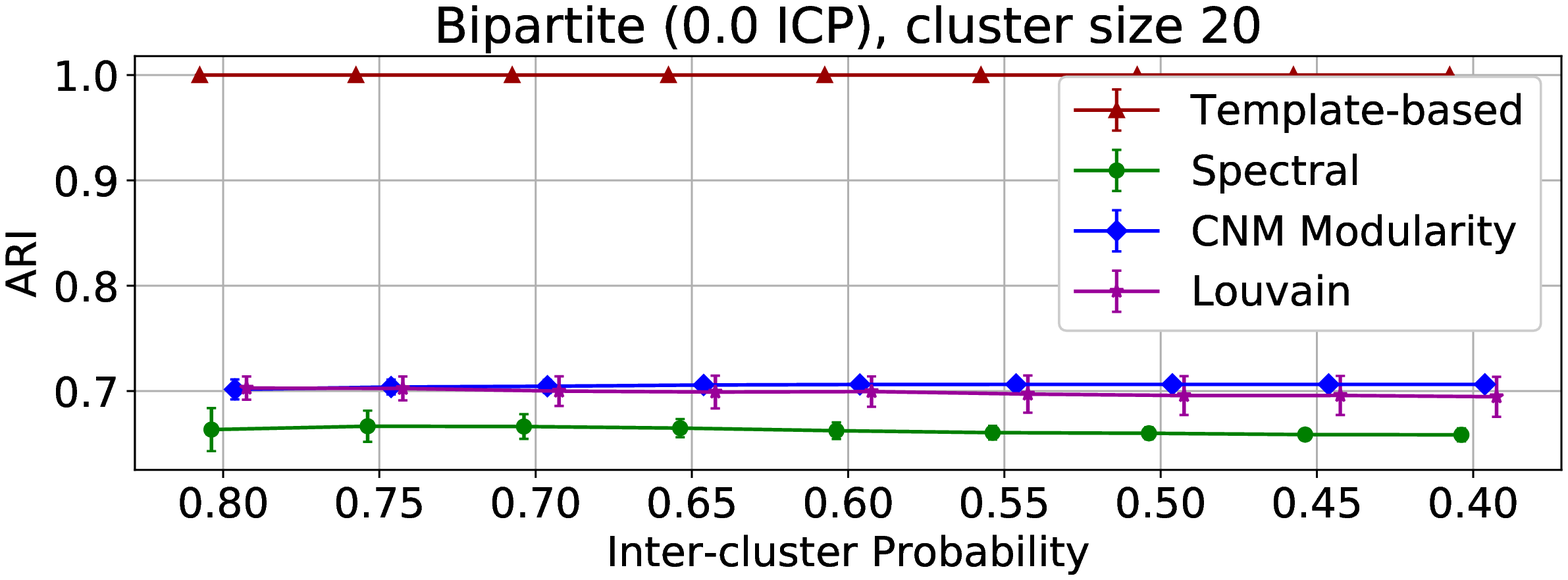} &
         \includegraphics[width=0.5\textwidth]{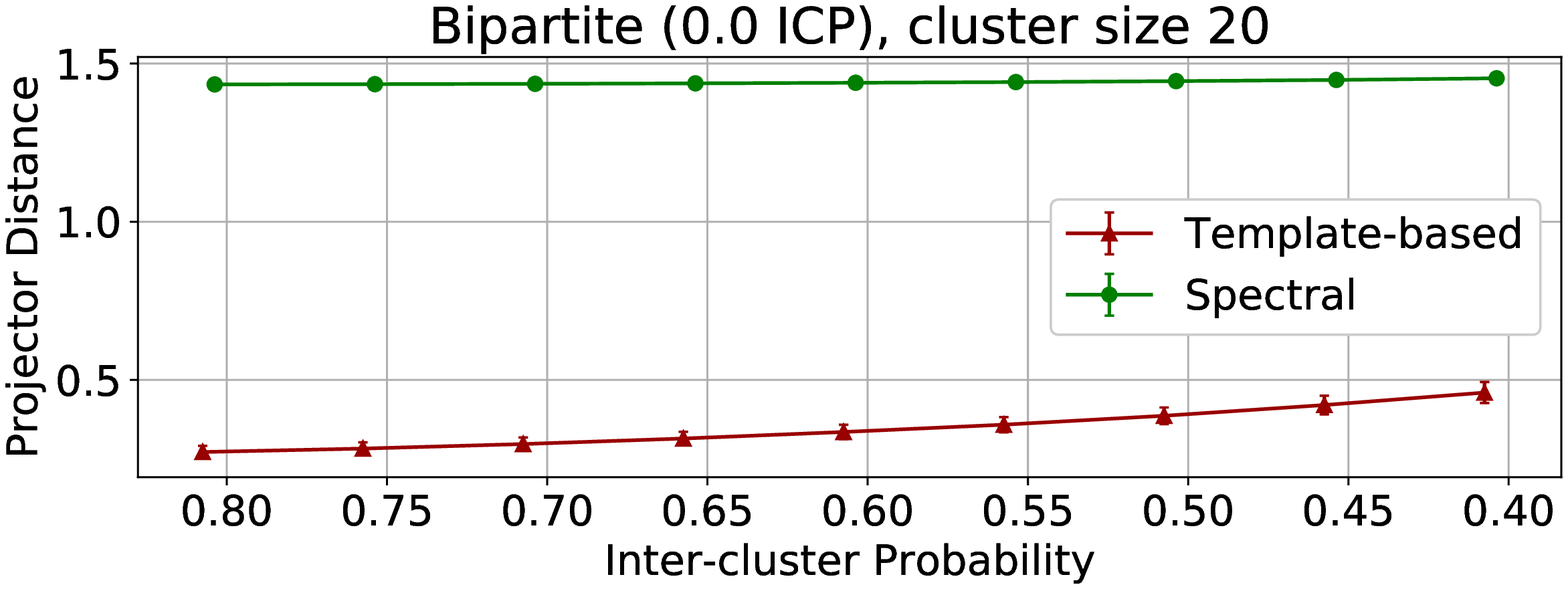} \\
         \includegraphics[width=0.5\textwidth]{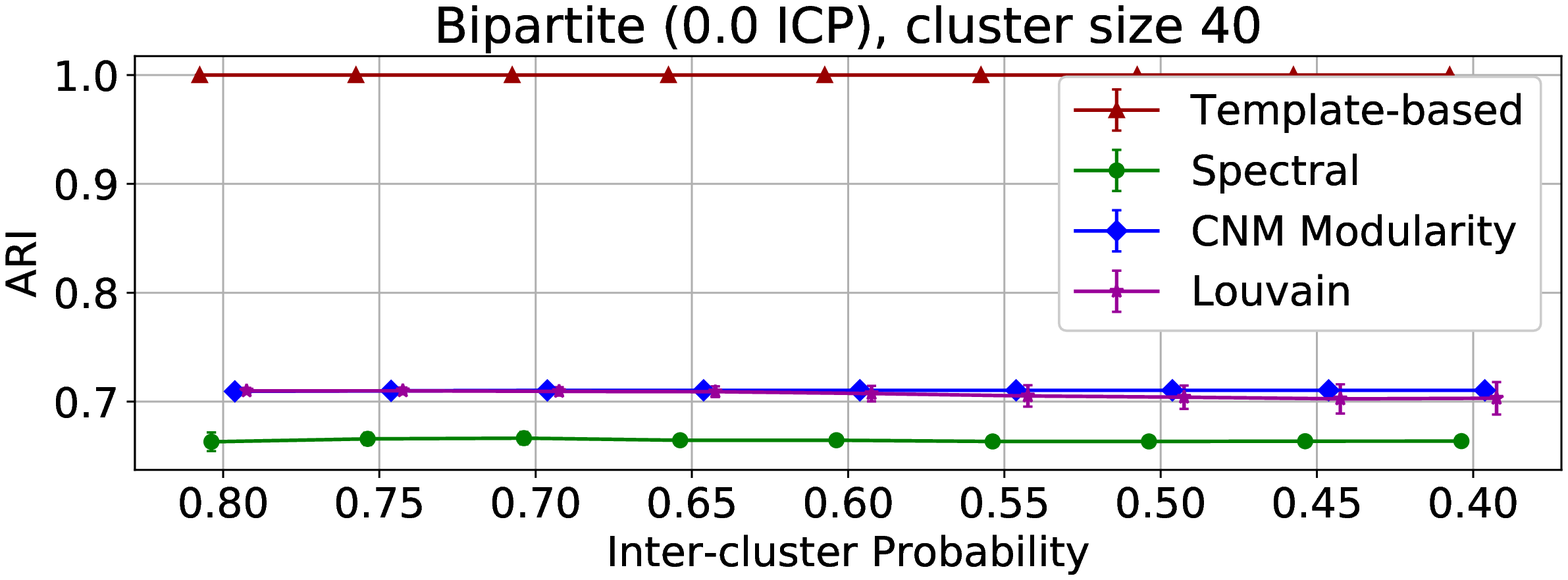} &
         \includegraphics[width=0.5\textwidth]{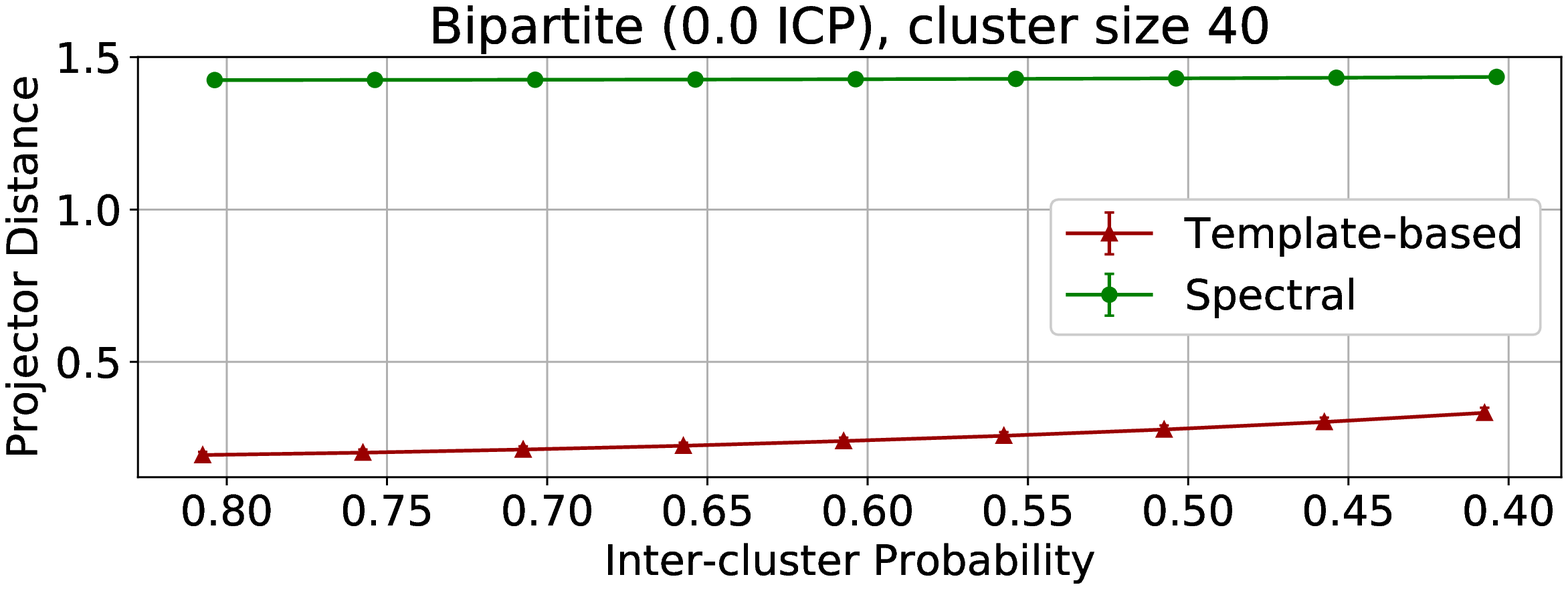}  \\
         \includegraphics[width=0.5\textwidth]{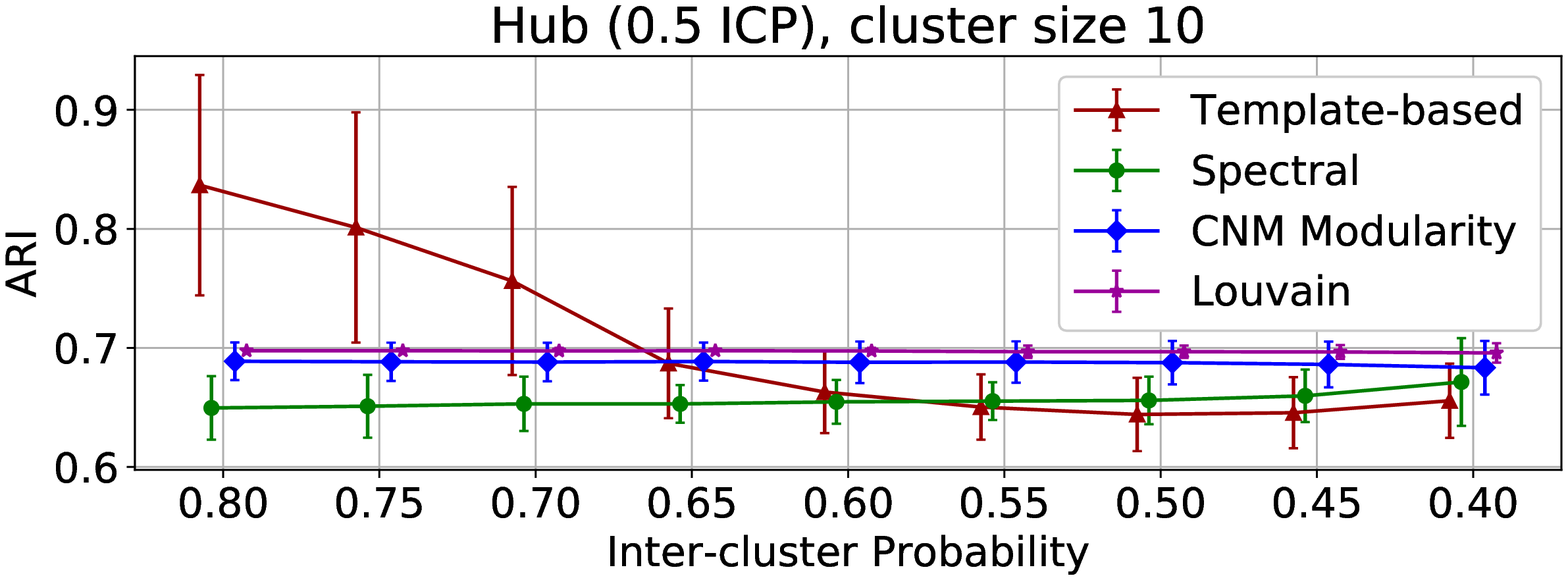} &
         \includegraphics[width=0.5\textwidth]{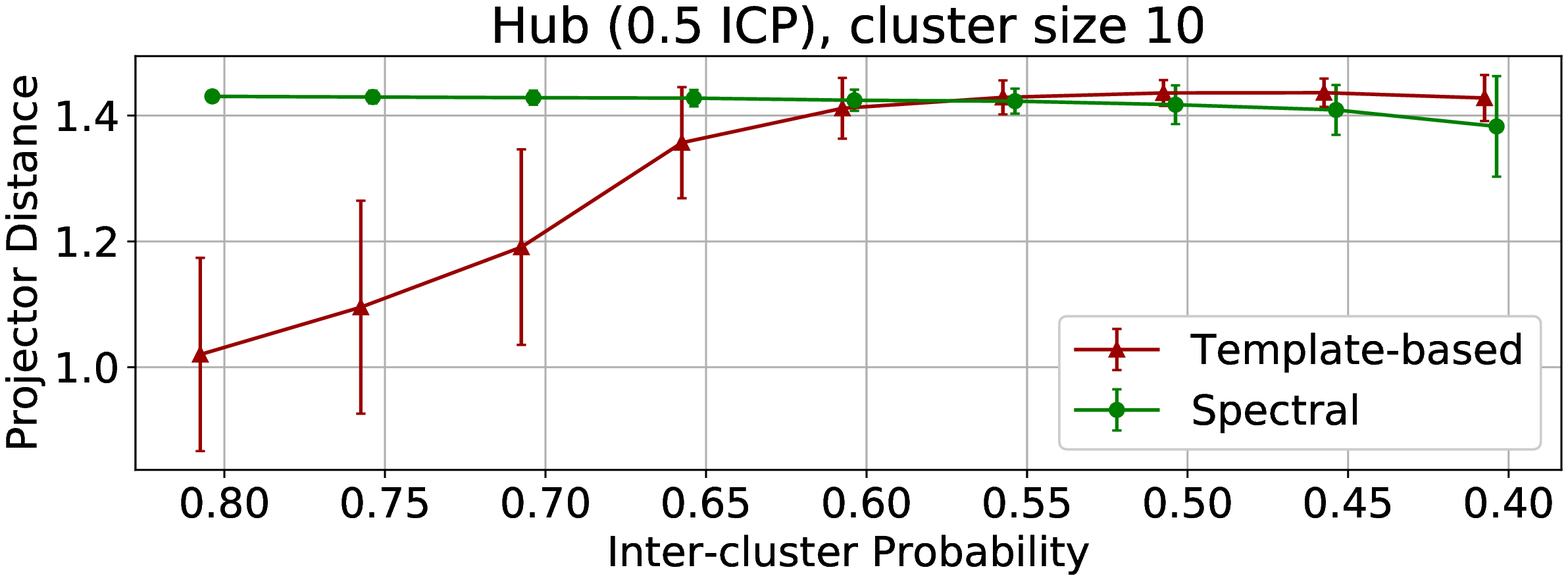} \\
         \includegraphics[width=0.5\textwidth]{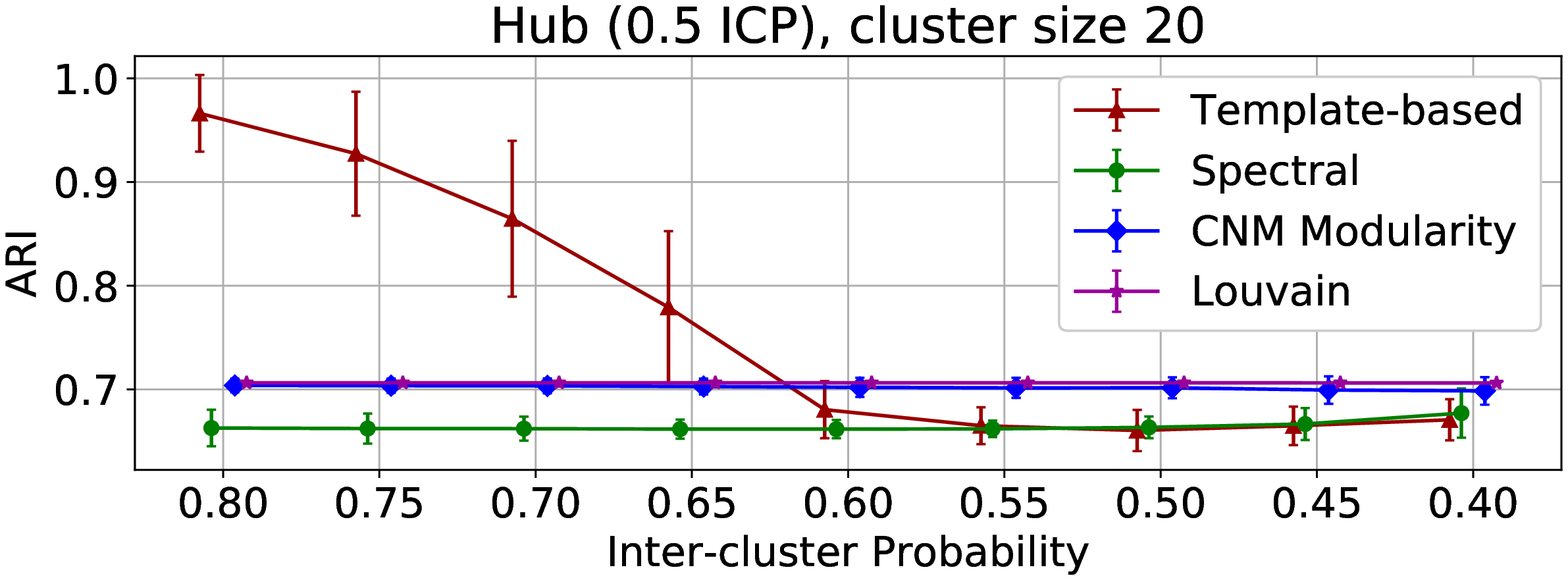} &
         \includegraphics[width=0.5\textwidth]{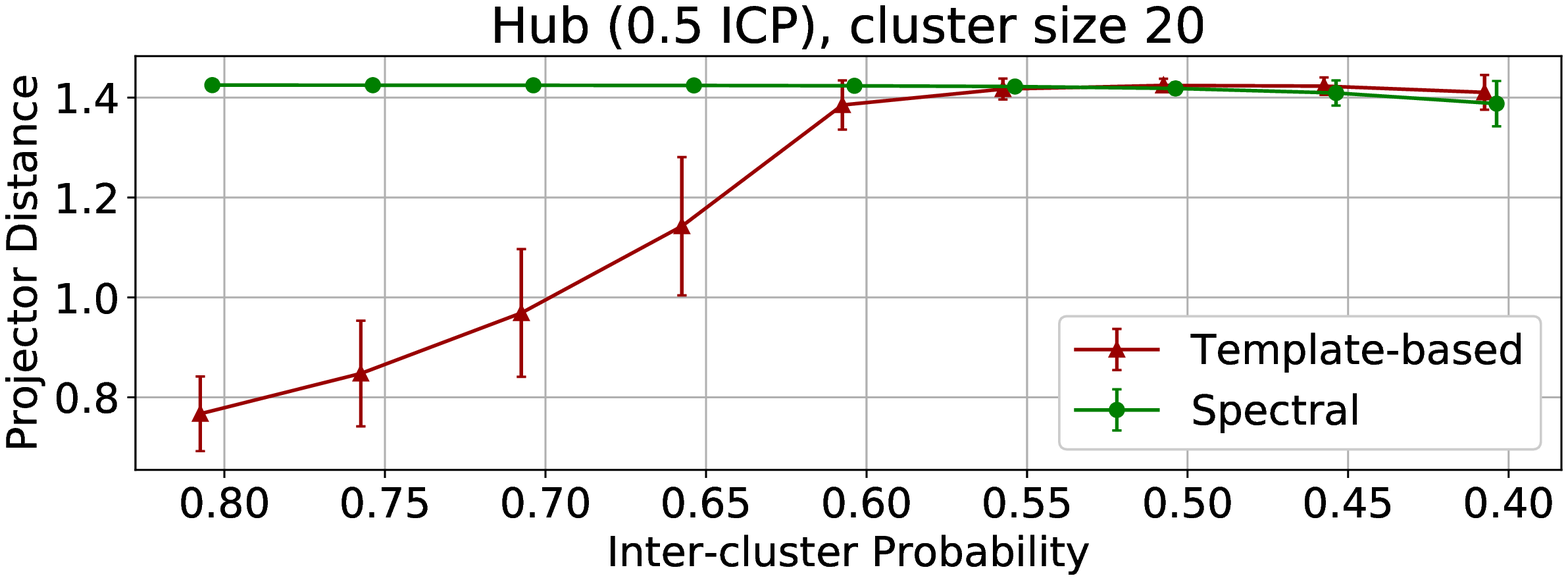} \\
         \includegraphics[width=0.5\textwidth]{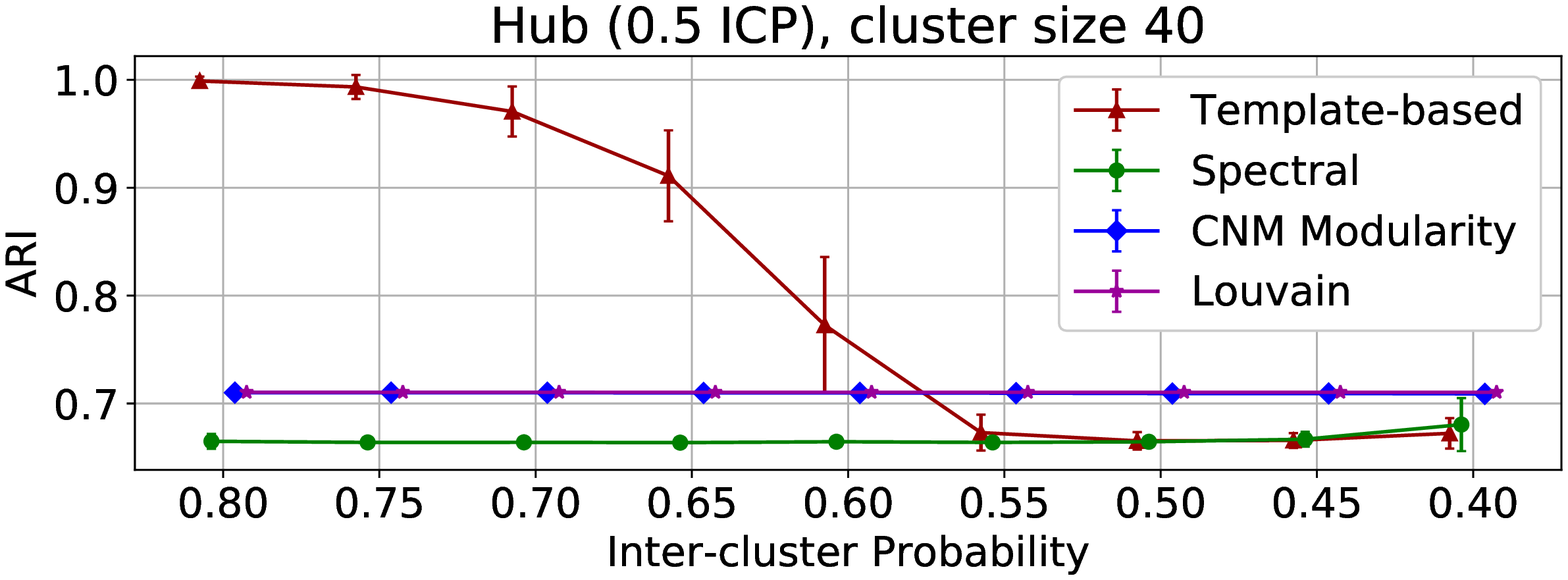} &
         \includegraphics[width=0.5\textwidth]{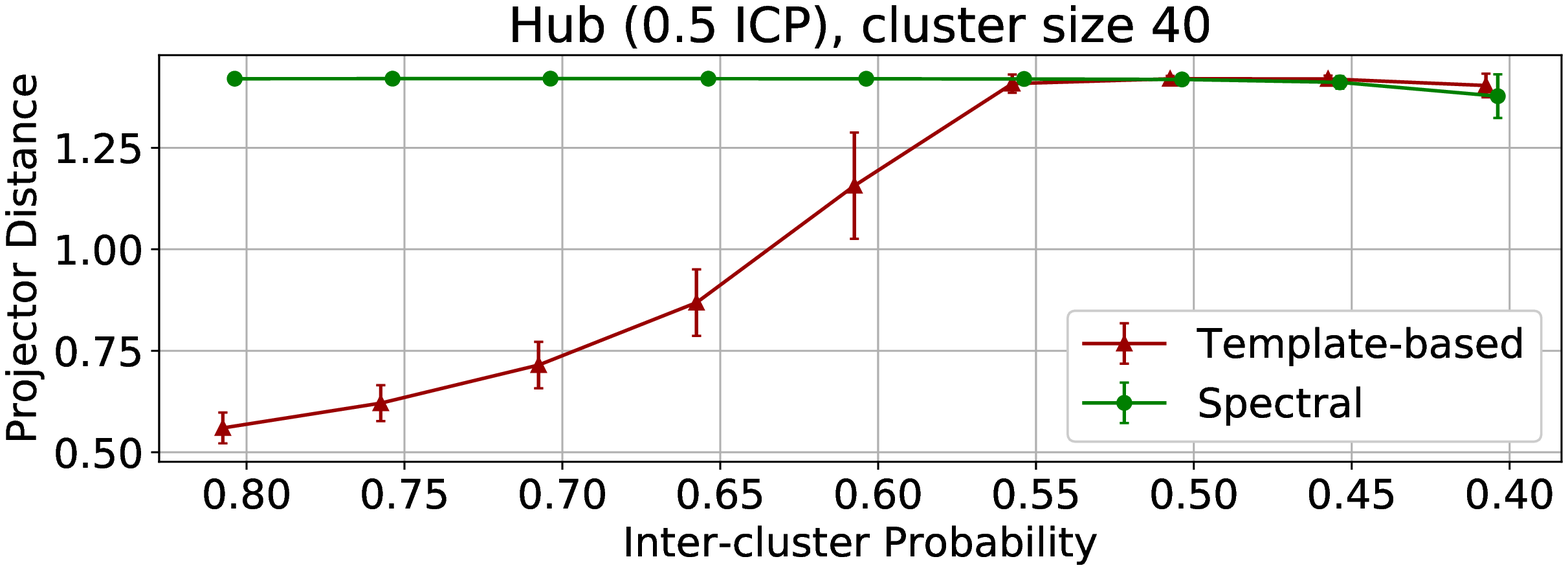}
    \end{tabular}
    \caption{Adjusted Rand Index and Projector Distance for the Bipartite and Hub Comparison experiments. Top three lines are for the ``Bipartite'' case, bottom three are for the ``Hub'' case. Error bars represent the standard deviation of the results. Rows correspond to different cluster sizes (10, 20, 40 for ``Bipartite'' and for ``Hub'').}
    \label{fig:bp_results}
\end{figure}

\emph{Bipartite and Hub Comparison Results:} Figure~\ref{fig:bp_results} shows the results for the Bipartite and Hub Comparison experiment.
As expected, both modularity and spectral graph clustering have greater difficulty clustering the bipartite cases, as their base versions are not equipped to deal with such graphs. By contrast, TB clustering performs accurately, with the ARI predictably falling off on the harder cases (such as a hub with few connections and small clusters, where many nodes are simply unconnected).

\subsection{Experiments on real datasets}
\label{ssec:real_experiments}
To validate the TB graph clustering method, we performed experiments on the \textit{email}\footnote{Taken from \url{https://snap.stanford.edu/data/email-Eu-core.html}} and the \textit{school}~\cite{Stehle2011-PONE} datasets.

The \textit{email} dataset is composed of $1\,005$ vertices, each representing an individual email account in a research institution. The graph has $25\,571$ edges, when emails are exchanged from one account to another; the average degree of a vertex is $33.8$ with a standard deviation of $37.4$. Each vertex belongs to one of 42 communities, depending on the department of the owner of the email account. 

The \textit{school} datasets are composed of $242$ vertices, each representing a student or teacher in a school. The \textit{school1} dataset has $37\,414$ edges and \textit{school2} has $40\,108$, representing interactions between individuals; average node degrees are $317$ and $338$ with standard deviations of $0.22$ and $0.27$ for each. Each vertex belongs to one of 11 communities, for each student and teacher's classes.

We used the previously annotated community ground truths to compute the models. To verify the robustness of our method to imperfections in the model, we also added Gaussian zero-mean noise to the weights in the models.
For the experiment, the following steps were taken:
\begin{enumerate}[topsep=0pt]
    \item Generate model graph from the ground truth communities, add noise with standard deviation $\sigma$;
    \item Execute TB, spectral, CNM modularity and Louvain modularity clustering;
    \item Evaluate the output of the clustering;
    \item Perform 40 repetitions with different random initializations, compute average and standard deviation of evaluation measures.
\end{enumerate}

\subsubsection{Results and Discussion.}
\label{sssec:real_results}

\begin{figure}[htbp]
    \centering
    \begin{tabular}{cc}
        \multicolumn{2}{c}{\textit{email}} \\
        \includegraphics[width=0.5\textwidth]{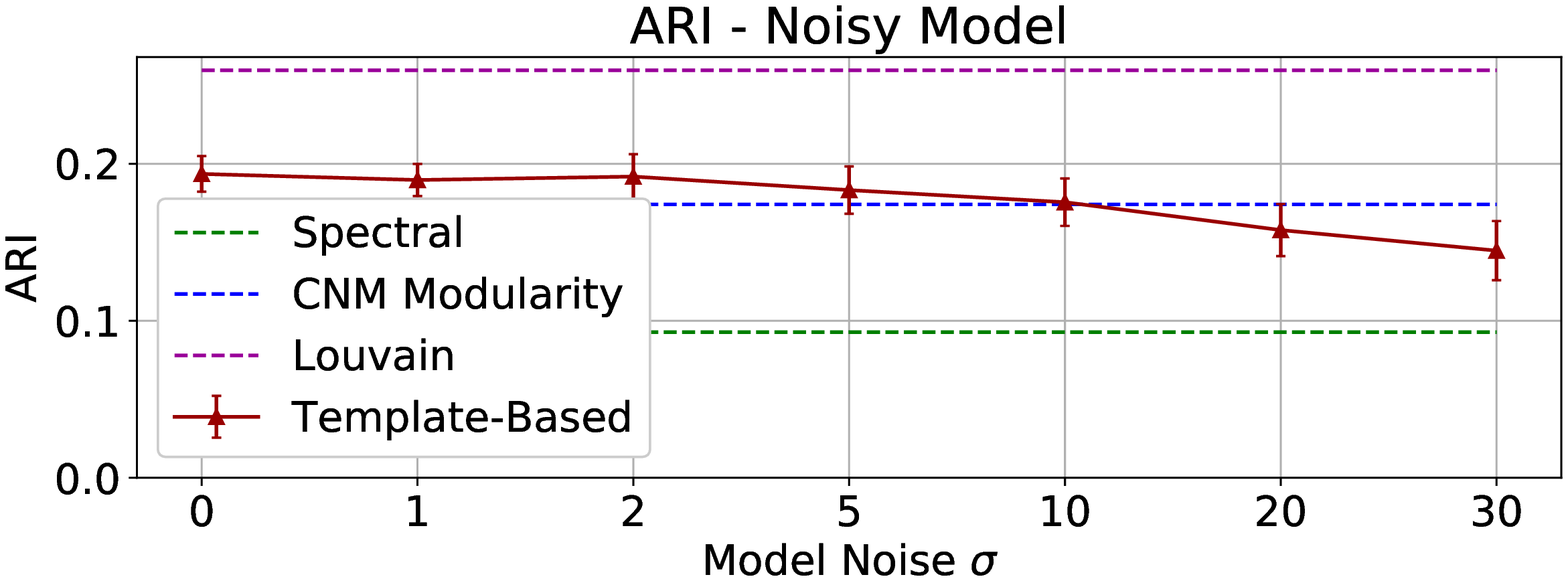} & 
        \includegraphics[width=0.5\textwidth]{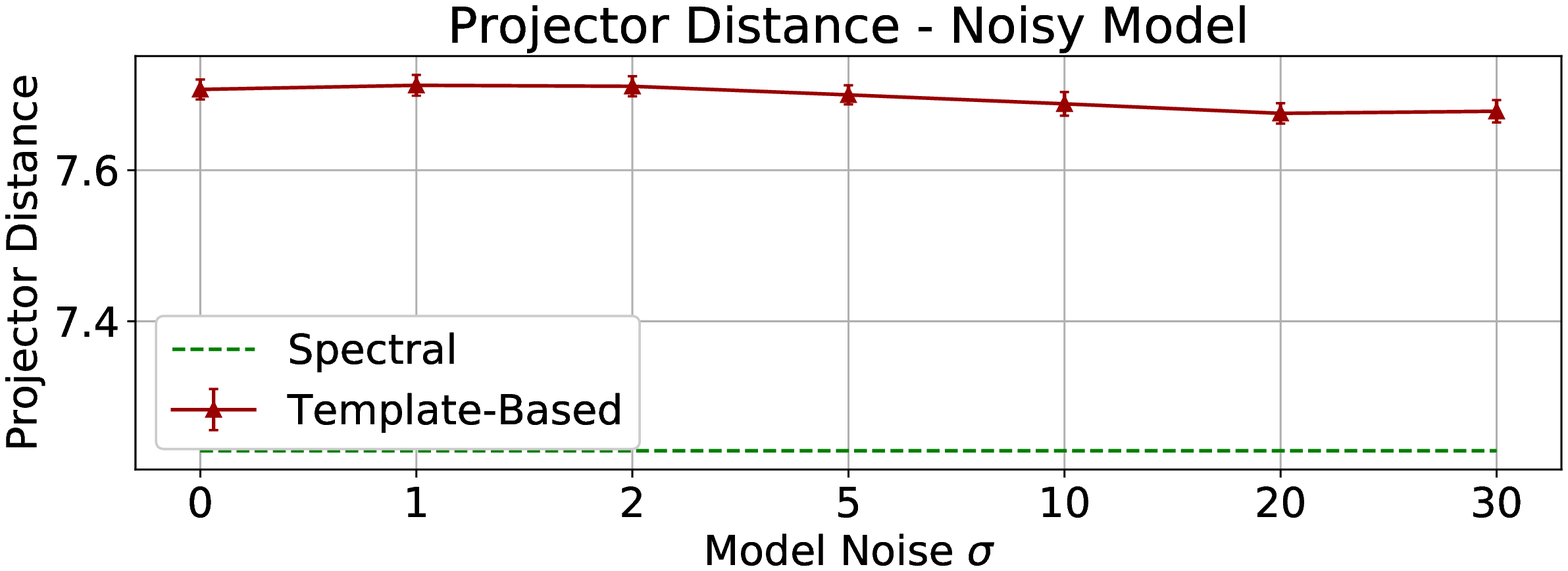} \\[3pt]
        \multicolumn{2}{c}{\textit{school1}} \\
        \includegraphics[width=0.5\textwidth]{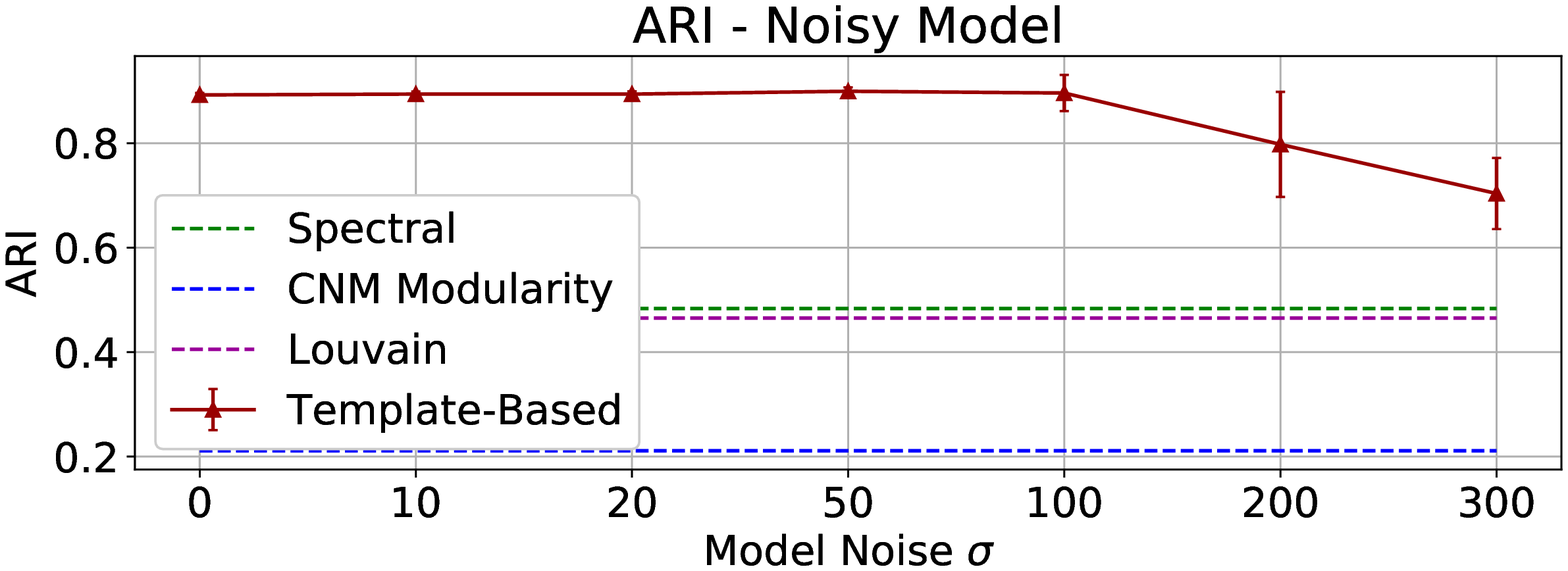} & 
        \includegraphics[width=0.5\textwidth]{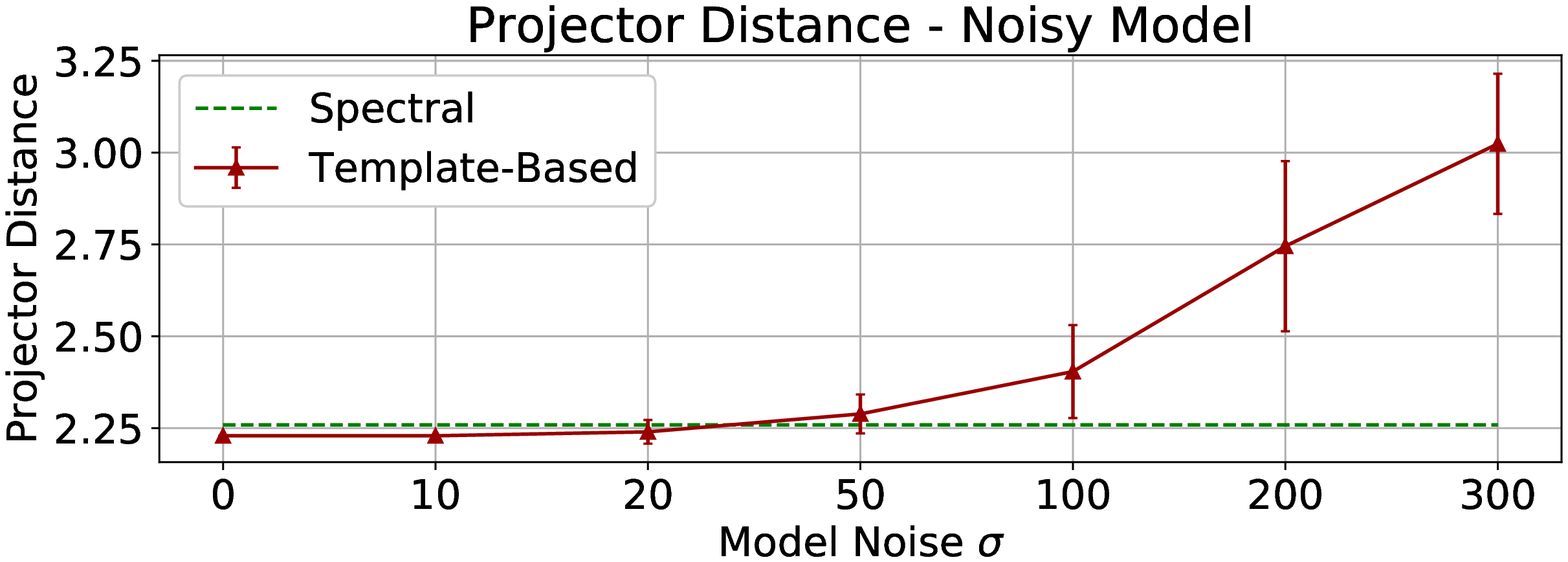} \\[3pt]
        \multicolumn{2}{c}{\textit{school2}} \\
        \includegraphics[width=0.5\textwidth]{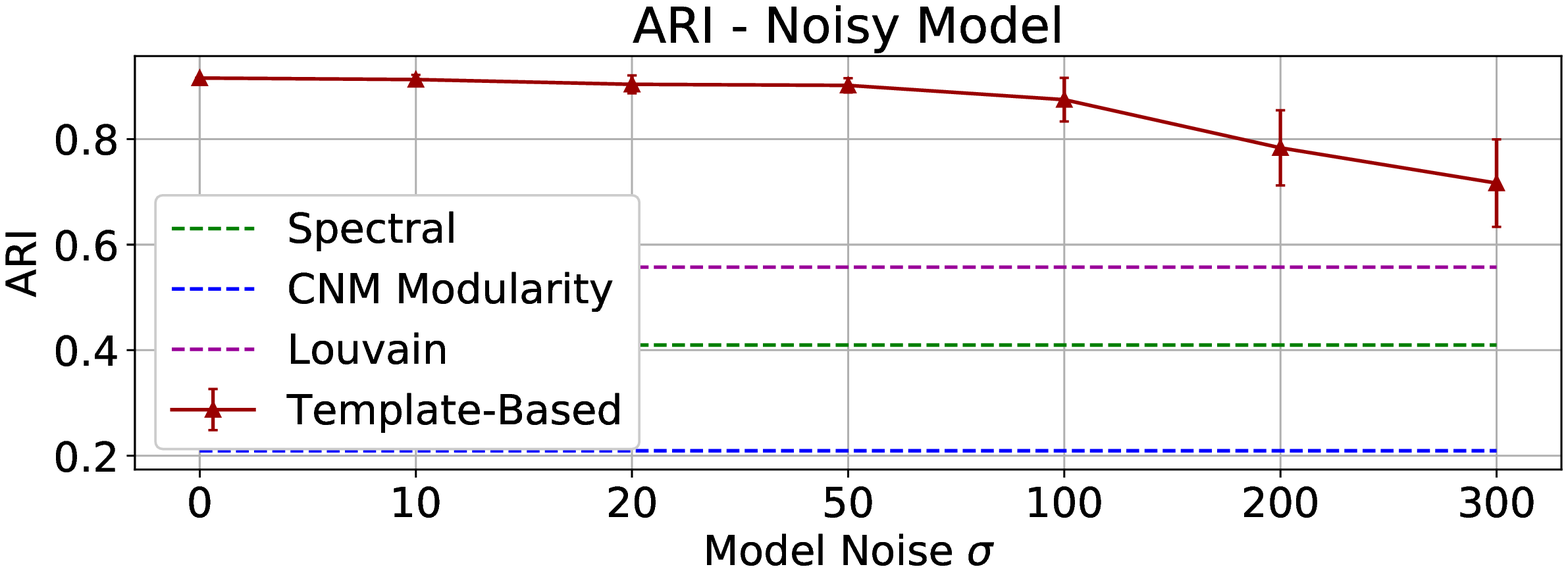} & 
        \includegraphics[width=0.5\textwidth]{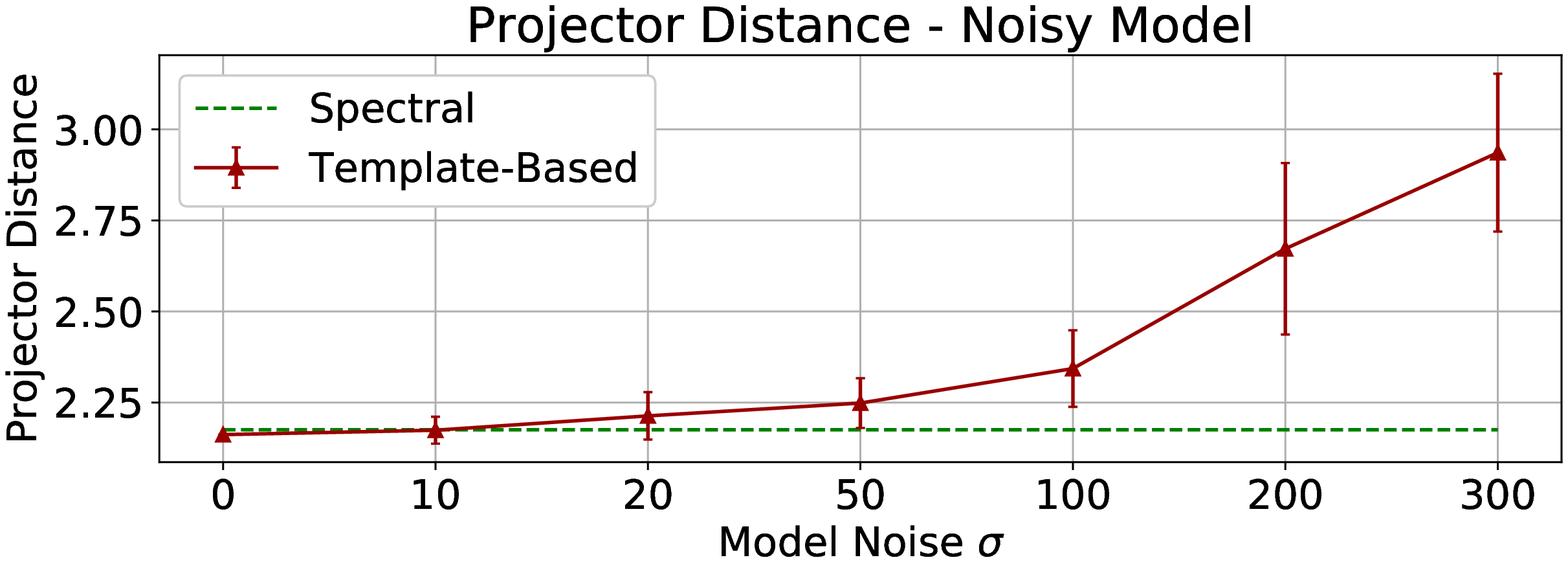} \\
    \end{tabular}
    \caption{Results for the experiment on the real datasets with noisy models, with $\sigma$ the standard deviation of the Gaussian noise added to each weight in the model. The dashed lines are the baseline methods (which are unaffected by noise).}
    \label{fig:noisy_results}
\end{figure}

Results for the real experiments are given in Table~\ref{tab:real_results} and Figure~\ref{fig:noisy_results}. We can see that the TB method can leverage the underlying structure of the communities to outperform CNM modularity and spectral clustering, while the low standard deviation shows that it is not significantly affected by different initializations, despite the intrinsic vulnerability of gradient descent methods to local minima. The Louvain modularity technique achieves a significantly better quality. In the \textit{email} dataset, the projector distance of the TB is slightly larger than the one of the spectral clustering; however, the spectral technique underperforms significantly compared to TB, pointing to an overall lack of quality of the spectral clustering, which may be caused by a failure of the $k$-means to properly segment the ``better'' embedding. In the \textit{school} datasets, however, TB largely outperforms all baselines, and edges out the spectral technique in the Projector Distance. This is justified by the leveraging of the strong information in our prior. Additionally, we can see that these observations hold even for moderate amounts of noise, demonstrating the robustness of the TB technique to imperfect models.

\begin{table}[htbp]
    \centering
    \caption{Results of the noiseless experiments on the real datasets: Adjusted Rand Index (ARI) and Projector Distance (PD).}
    \label{tab:real_results}
    \begin{tabular}{c|c|cccc}
Dataset name & Metric & TB & Spectral & CNM Modularity & Louvain\\
\hline
\hline
\multirow{3}{*}[5pt]{email} & ARI & $0.19 \pm 0.01$ & $0.10 \pm 0.03$ & $0.17 \pm 0.0$ & $\mathbf{0.26 \pm 0.0}$\\
 & Proj. Dist. & $7.71 \pm 0.01$ & $\mathbf{7.23 \pm 0.0}$ & N/A & N/A\\
 \hline
\multirow{3}{*}[5pt]{school1} & ARI & $\mathbf{0.89 \pm 0.0}$ & $0.47 \pm 0.03$ & $0.21 \pm 0.0$ & $0.47 \pm 0.0$\\
 & Proj. Dist. & $\mathbf{2.23 \pm 0.0}$ & $2.26 \pm 0.0$ & N/A & N/A\\
 \hline
\multirow{3}{*}[5pt]{school2} & ARI & $\mathbf{0.92 \pm 0.0}$ & $0.41 \pm 0.02$ & $0.21 \pm 0.0$ & $0.56 \pm 0.0$\\
 & Proj. Dist. & $\mathbf{2.16 \pm 0.0}$ & $2.18 \pm 0.0$ & N/A & N/A
    \end{tabular}
\end{table}

\section{Conclusion}
\label{sec:conclusion}

In this paper, we presented a novel method for clustering a graph when we have prior information on the structure of its communities. 
Although this is a strong prior and not readily available for many cases, we believe that this method is a significant step towards improving the quality of clustering and segmentation of data containing this type of prior, such as graphs derived from images of structured scenes.
Our results show the potential of the template-based method for clustering graphs with known prior about their community structure. We obtained equal or greater accuracy than common methods, such as spectral clustering or modularity, and in particular for scenarios where communities have little or no internal connections.

There are many possible avenues of research for extending the template-based graph clustering technique. In several real-world applications, nodes of a network have multiple underlying communities, e.g. in a product recommendation network, each product may belong to several categories. It could be modified to support multi-labeled nodes. Further experiments on real datasets and applications, such as graphs extracted from medical images\cite{Caetano2009-PAMI}, should also be performed, to confirm and explore the advantages and limitations of the technique. Improvements on the algorithm complexity, as well as faster implementations, can result in a speed-up of the technique. Finally, and perhaps of greater interest, a theoretical approximation of the spectral graph theory with the template-based clustering could be performed by searching for a graph Laplacian-based model and matching algorithm.

%
%

\bibliographystyle{splncs04}
\bibliography{refs}
\end{document}